\documentclass[10pt,twocolumn,letterpaper]{article}

\usepackage[pagenumbers]{cvpr} 

\def\cvprPaperID{6872} 
\def\confName{CVPR}
\def\confYear{2022}

\usepackage{graphicx}
\usepackage{amsmath}
\usepackage{amssymb}
\usepackage{booktabs}

\usepackage{dsfont}
\usepackage{multirow}
\usepackage{kotex}
\usepackage[dvipsnames]{xcolor}
\usepackage{algorithm}
\usepackage{listings}

\usepackage{calc}
\makeatletter
\let\latex@fnsymbol\@fnsymbol
\renewcommand\@fnsymbol[1]{\ifcase#1\or$\dagger$\or*\else\@ctrerr\fi}
\newcommand{\restorefnsymbol}{\let\@fnsymbol\latex@fnsymbol}
\makeatother

\usepackage[pagebackref,breaklinks,colorlinks]{hyperref}

\usepackage{enumitem}

\usepackage[capitalize]{cleveref}
\crefname{section}{Sec.}{Secs.}
\Crefname{section}{Section}{Sections}
\Crefname{table}{Table}{Tables}
\crefname{table}{Tab.}{Tabs.}

\begin{document}

\title{Feature Statistics Mixing Regularization for Generative Adversarial Networks}

\author{Junho Kim\textsuperscript{1} \thinspace\quad Yunjey Choi\textsuperscript{1} \thinspace\quad Youngjung Uh\textsuperscript{2}\thanks{Corresponding author.} \\
\\
\textsuperscript{1}NAVER AI Lab \quad  \textsuperscript{2}Yonsei University
}


\maketitle
\definecolor{olive}{rgb}{0.5, 0.5, 0.0}
\definecolor{maroon}{rgb}{0.69, 0.19, 0.38}
\definecolor{celestialblue}{rgb}{0.29, 0.59, 0.82}
\definecolor{darkgreen}{rgb}{0.0, 0.5, 0.0}
\definecolor{grey}{rgb}{0.5,0.5,0.5}
\definecolor{darkblue}{rgb}{0.19, 0.19, 0.62}
\definecolor{nicergreen}{rgb}{0.13, 0.54, 0.13}
\definecolor{nicered}{rgb}{0.83, 0.16, 0.16}
\newcommand{\std}[1]{\makebox[6.3mm][r]{\scriptsize\raisebox{0.25mm}{\scalebox{0.7}{$\pm$}}\hspace{0.2mm}#1}}
\newcommand{\fref}[1]{Figure \ref{#1}}
\newcommand{\eref}[1]{Eq. \ref{#1}}
\newcommand{\tref}[1]{Table \ref{#1}}
\newcommand{\sref}[1]{Section \ref{#1}}

\newcommand{\TODO}[1]{{\color{red}TODO: #1}}
\newcommand{\remove}[1]{{\color{BrickRed}#1}}
\newcommand{\uh}[1]{{\color{violet}#1}}
\newcommand{\jh}[1]{{\color{RoyalPurple}#1}}
\newcommand{\yjc}[1]{{\color{teal}#1}}

\newcommand{\FINAL}[1]{#1}

\def\clap#1{\hbox to 0pt{\hss #1\hss}}%
\def\initials#1{\protect\clap{\smash{\raisebox{1.4ex}{\tiny{\textsf{\textit{#1}}}}}}}%
\newcommand{\EDIT}[4][]{\strut{\color{#3}{\hspace{0pt}\initials{#2}{\color{red}\sout{#1}}{#4}}}}
\newcommand{\TK}[2][]{\protect\EDIT[#1]{TK}{olive}{#2}}
\newcommand{\TA}[2][]{\protect\EDIT[#1]{TA}{maroon}{#2}}
\newcommand{\JL}[2][]{\protect\EDIT[#1]{JL}{celestialblue}{#2}}
\newcommand{\SL}[2][]{\protect\EDIT[#1]{SL}{olive}{#2}}
\newcommand{\MA}[2][]{\protect\EDIT[#1]{MA}{darkblue}{#2}}

\newcommand{\norm}[1]{\left\lVert#1\right\rVert}
\newcommand{\abs}[1]{\lvert#1\rvert}
\newcommand{\real}{\mathbb{R}}
\newcommand{\integer}{\mathbb{Z}}
\newcommand{\proby}{\mathbf{y}}
\newcommand{\probx}{\mathbf{x}}
\newcommand{\probz}{\mathbf{z}}
\newcommand{\probn}{\mathbf{n}}
\newcommand{\aug}{\mathcal{T}}
\newcommand{\augu}{\mathcal{U}}
\newcommand{\augr}{\mathcal{R}}
\newcommand{\augg}{\mathcal{G}}
\newcommand{\augp}{\mathcal{P}}
\newcommand{\augq}{\mathcal{Q}}
\newcommand{\augeye}{\mathcal{I}}

\newcommand{\bmc}{\mathbf{c}\xspace}
\newcommand{\bms}{\mathbf{s}\xspace}
\newcommand{\bmx}{\mathbf{x}\xspace}
\newcommand{\bmy}{\mathbf{y}\xspace}
\newcommand{\bmxt}{\mathbf{y}\xspace}
\newcommand{\bmf}{{f}\xspace}
\newcommand{\bmD}{D\xspace}
\newcommand{\adain}{\text{AdaIN}\xspace}
\newcommand{\mixup}{\text{Mixup}\xspace}
\newcommand{\fsm}{\text{FSM}\xspace}
\newcommand{\linear}{\text{Linear}\xspace}
\newcommand{\bmhc}{{\bmh}_{c}\xspace}
\newcommand{\bmhs}{{\bmh}_{s}\xspace}
\newcommand{\bmh}{\mathbf{h}\xspace}
\newcommand{\bmhx}{{\bmx}\xspace}
\newcommand{\bmhxt}{{\bmxt}\xspace}
\newcommand{\bmhy}{{\bmh}_{\bmy}\xspace}
\newcommand{\bmmu}{\mathbf{\mu}\xspace}
\newcommand{\bmsigma}{\mathbf{\sigma}\xspace}

\newcommand{\h}{0mm}
\newcommand{\hh}{0mm}
\newcommand{\hhh}{0mm}
\newcommand{\hhhh}{0mm}
\newcommand{\vv}{0mm}
\newcommand{\vvv}{0mm}
\newcommand{\vvvv}{0mm}
\newcommand{\vvvvv}{0mm}
\newcommand{\s}{\hphantom{0}}
\newcommand{\cs}{\hspace{0}}
\newcommand{\thinfbox}[1]{\begingroup\setlength{\fboxrule}{0.01mm}\setlength{\fboxsep}{-\fboxrule}\fbox{#1}\endgroup}
\newcommand{\placeholderfig}[3]{\raisebox{\fboxsep+\fboxrule}{\fbox{\parbox[b][#2-2\fboxsep-2\fboxrule][t]{#1-2\fboxsep-2\fboxrule}{\center #3}}}} %
\newcommand{\subtabletop}[4]{\parbox[b][#2]{#1}{\footnotesize\centering\scalebox{#3}{#4}\vfill}} %
\newcommand{\subtablemid}[4]{\parbox[b][#2]{#1}{\footnotesize\centering\vfill\scalebox{#3}{#4}\vfill}} %
\newcommand{\subtablebot}[4]{\parbox[b][#2]{#1}{\footnotesize\centering\vfill\scalebox{#3}{#4}\vspace{0mm}}} %

\def\methodshort{FSMR\xspace}
\newrobustcmd{\B}{\bfseries}
\newcommand{\figsensitivity}{
\begin{figure}[t]
\centering
\includegraphics[width=1.0\linewidth]{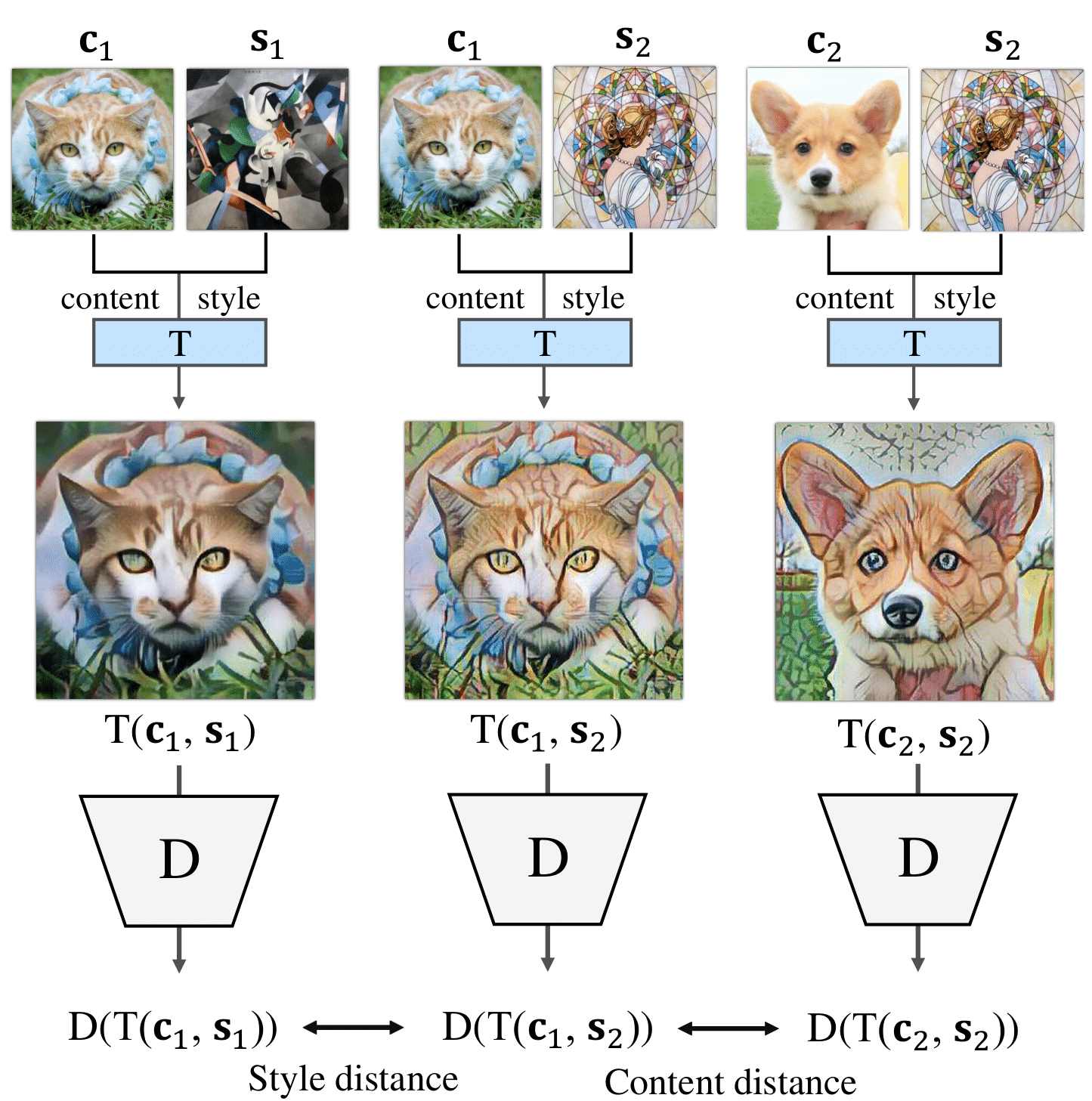}
\makebox{(a) Style distance and content distance}
\includegraphics[width=1.0\linewidth]{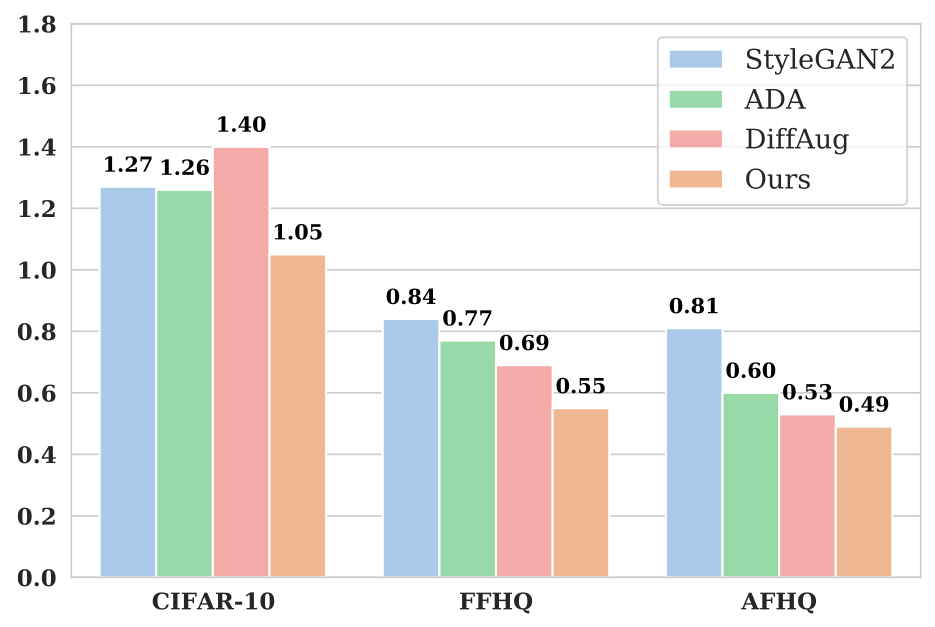}
\makebox{(b) Relative distance}
\caption{
(a) The style transfer method $T(\bmc, \bms)$ transfers the style of $\bms$ on the content of $\bmc$. We define style distance as the output difference due to style variations. Content distance is defined vice versa. 
(b) Relative distance across various GAN methods. Relative distance indicates how sensitive a discriminator is to style changes (\eref{eq:rho}).
See Section \ref{sec:style_bias} for details.
} 
\label{fig:fig1}
\end{figure}
}

\newcommand{\figmodel}{
\begin{figure*}[t]
\centering
\includegraphics[width=1.0\linewidth]{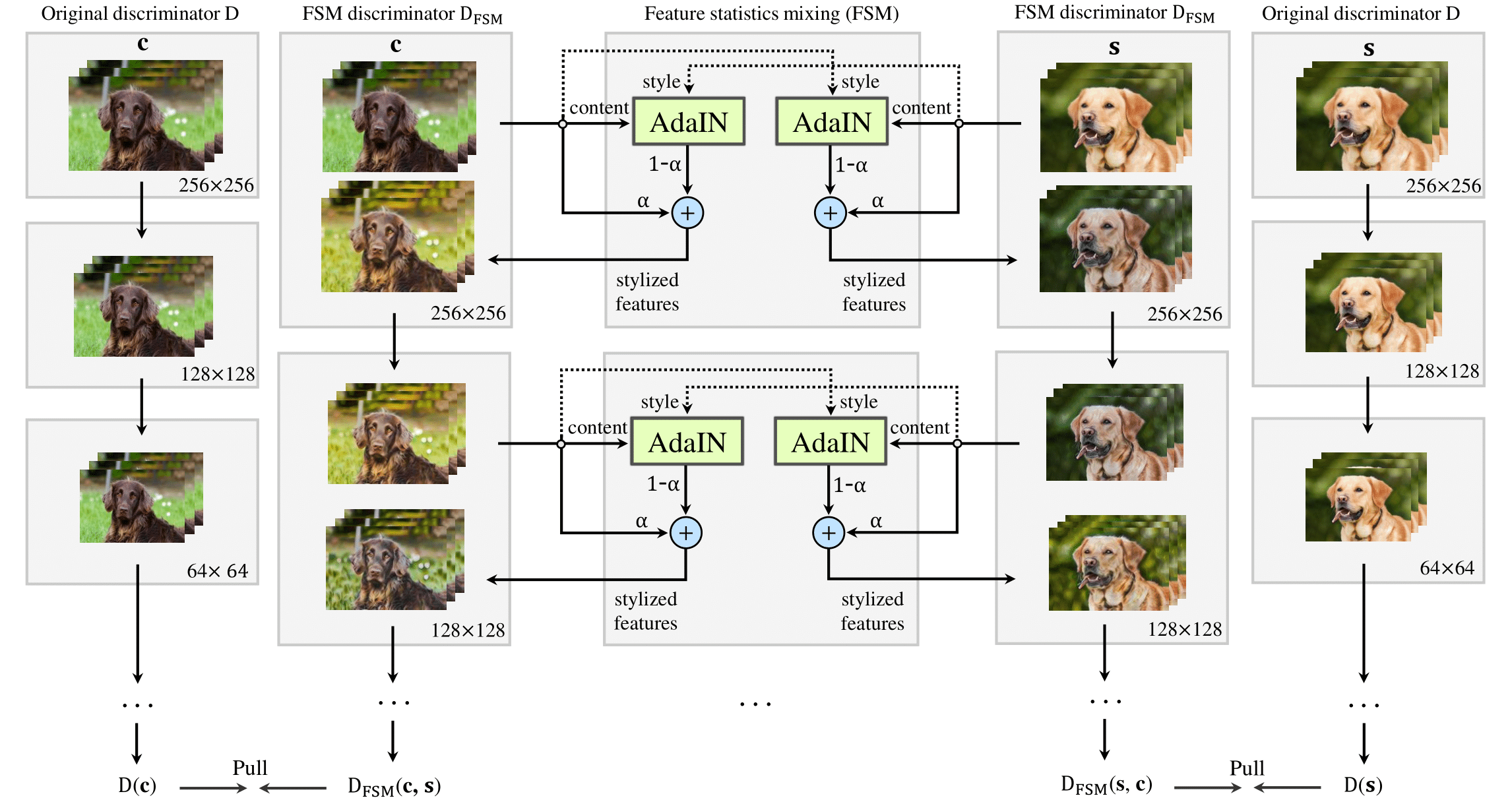}\\
\caption{
\textbf{Overview of feature statistics mixing regularization} (\sref{sec:fsmr}). Within the forward pass in the discriminator, we perturb features by applying AdaIN with a different sample. In deeper layers, the perturbations are applied recursively. A scalar $\alpha \sim \text{Uniform}(0, 1)$ moderates their strength. Then we enforce similarity between the original output and the perturbed one.
}
\label{fig:fig2}
\end{figure*}
}

\newcommand{\figrecon}{
\begin{figure*}[t]
\centering
\newcommand{\ww}{0.07\linewidth}%
\newcommand{\w}{0.28\linewidth}%
\renewcommand{\h}{0.21\linewidth}%

\includegraphics[width=\ww,height=\h]{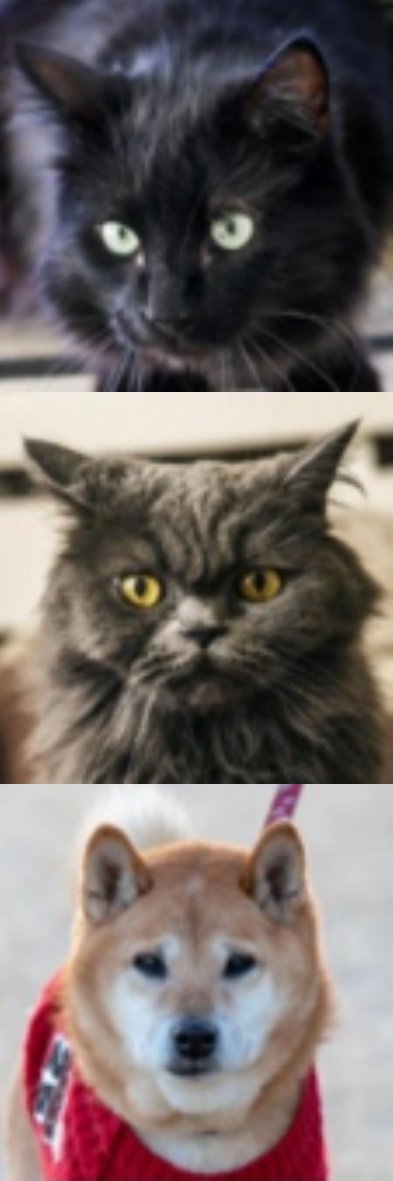}\hfill%
\hspace{0.1mm}
\includegraphics[width=\w,height=\h]{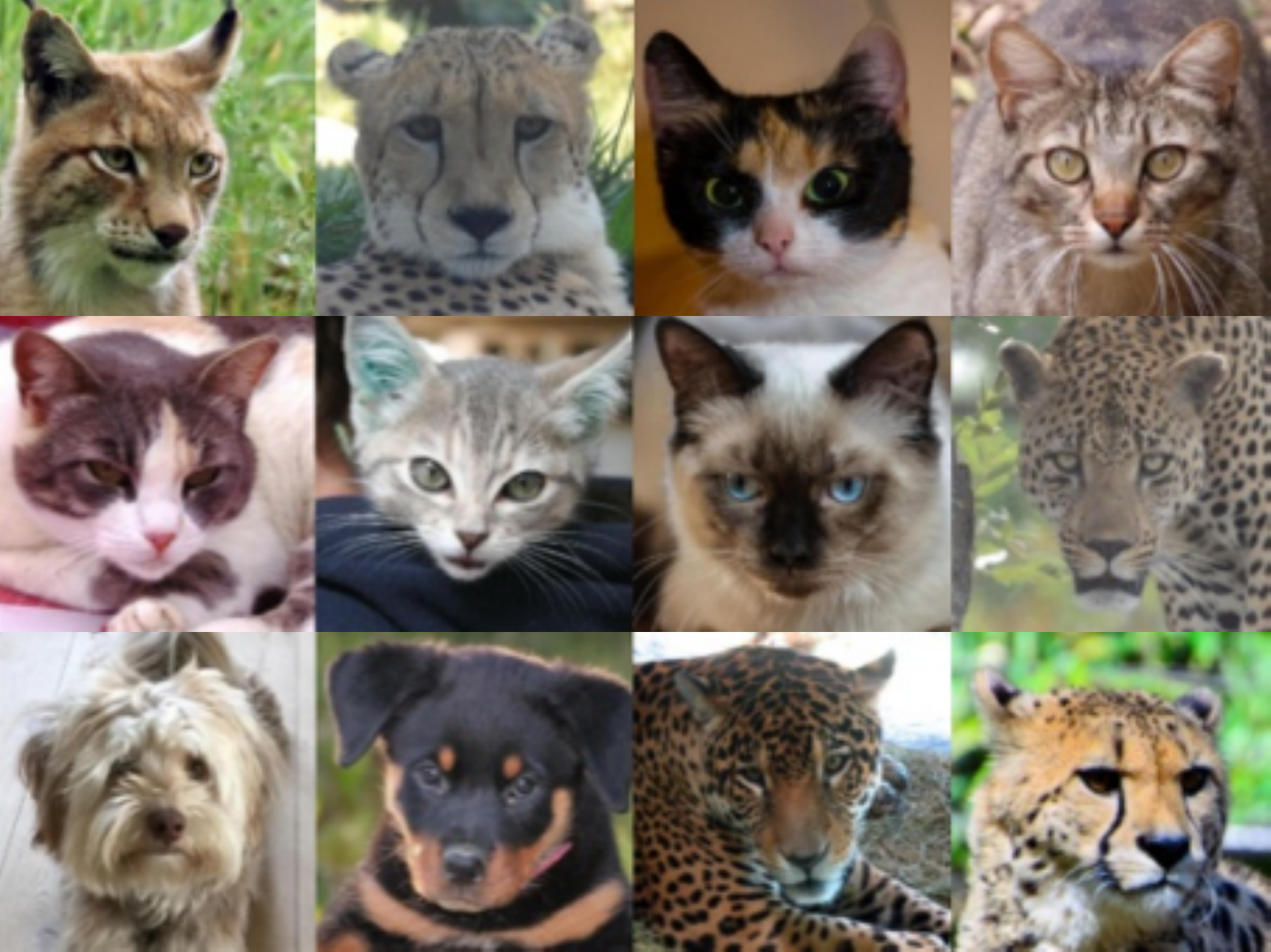}\hfill%
\hspace{0.1mm}
\includegraphics[width=\w,height=\h]{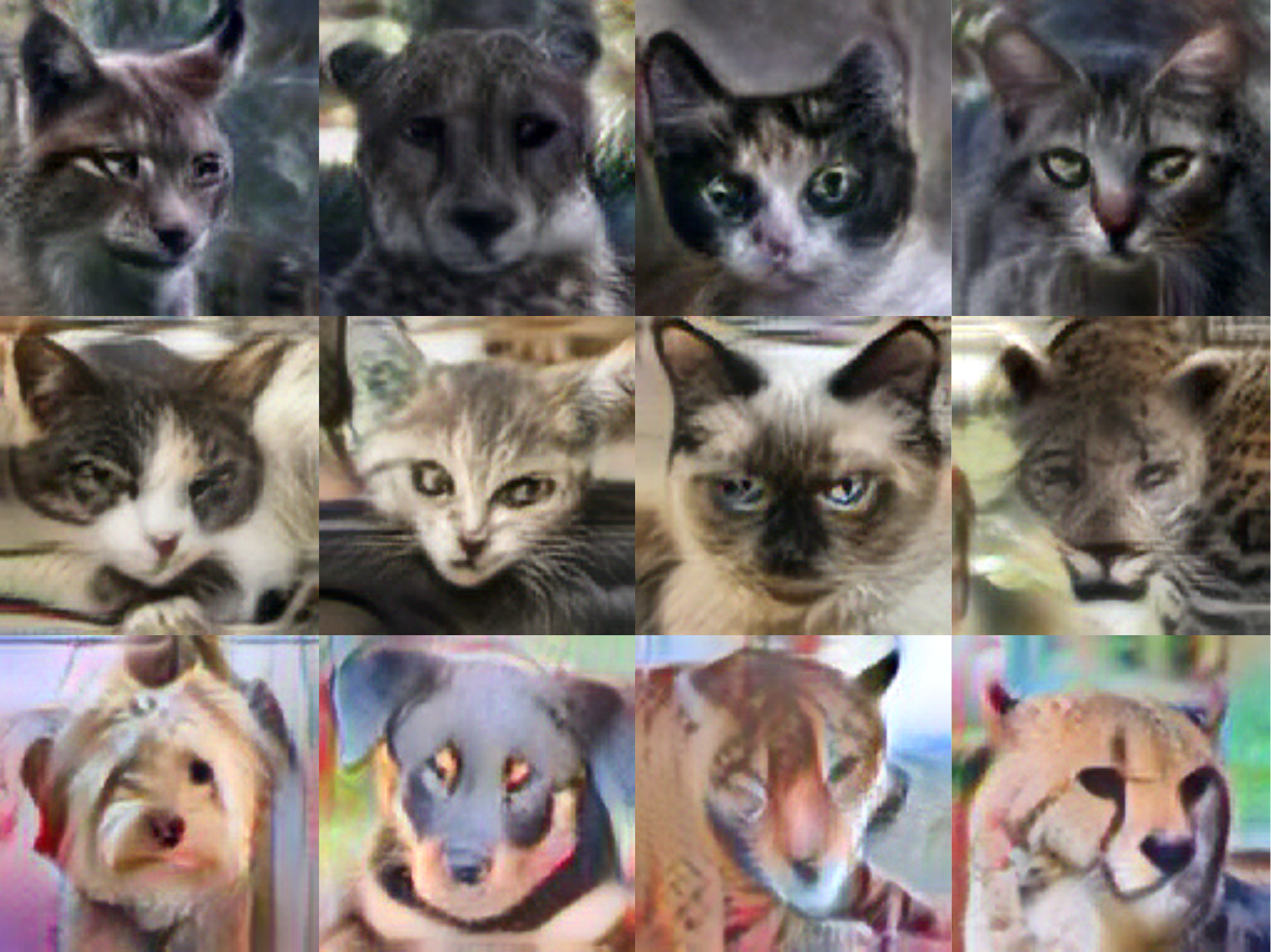}
\hfill%
\hspace{0.1mm}
\includegraphics[width=\w,height=\h]{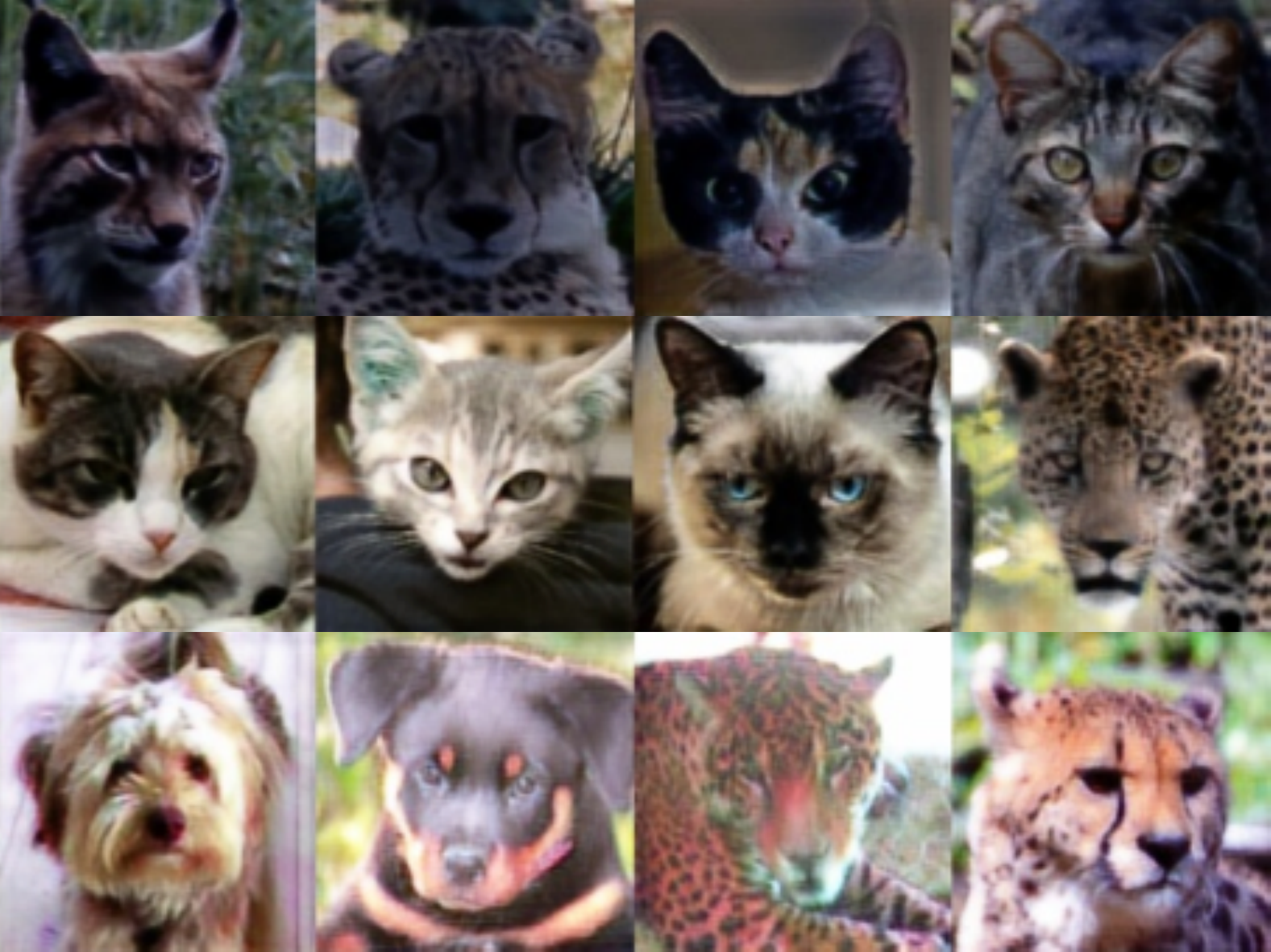}\hfill%

{\small
\makebox[\ww][c]{(a) Style}\hfill%
\hspace{0.1mm}
\makebox[\w][c]{(b) Content}\hfill%
\hspace{0.1mm}
\makebox[\w][c]{(c) Stylization by AdaIN \cite{huang2017arbitrary}}\hfill%
\hspace{0.1mm}
\makebox[\w][c]{(d) Visualization of FSM}\hfill%
}

\caption{
\textbf{Visualization of the effect of FSM} (\sref{sec:visualization}). 
(a) Example style images. (b) Example content images. (c) AdaIN largely distorts fine details. (d) Reconstruction of FSMed features preserves them. 
}
\label{fig:fig3}
\end{figure*}
}

\newcommand{\figensitivityresult}{
\begin{figure*}[t]
\begin{center}

\includegraphics[width=1.0\linewidth]{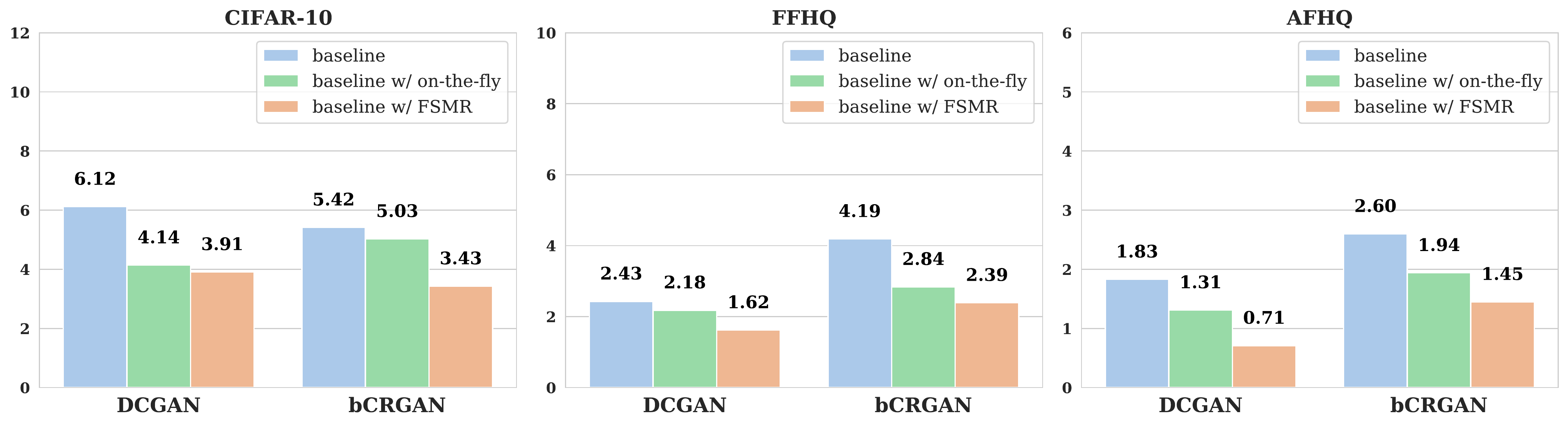}

\end{center}
\vspace{-6pt}
\caption{
\textbf{The relative distance of the discriminators} on CIFAR-10, FFHQ, and AFHQ. We observe a positive correlation with FID in each case. See Appendix \ref{app:addresults} for more results on other baselines and datasets.
}
\label{fig:fig4}
\end{figure*}
}

\newcommand{\figCIFAR}{
\begin{figure*}[t]
\centering%

\includegraphics[width=1.0\linewidth]{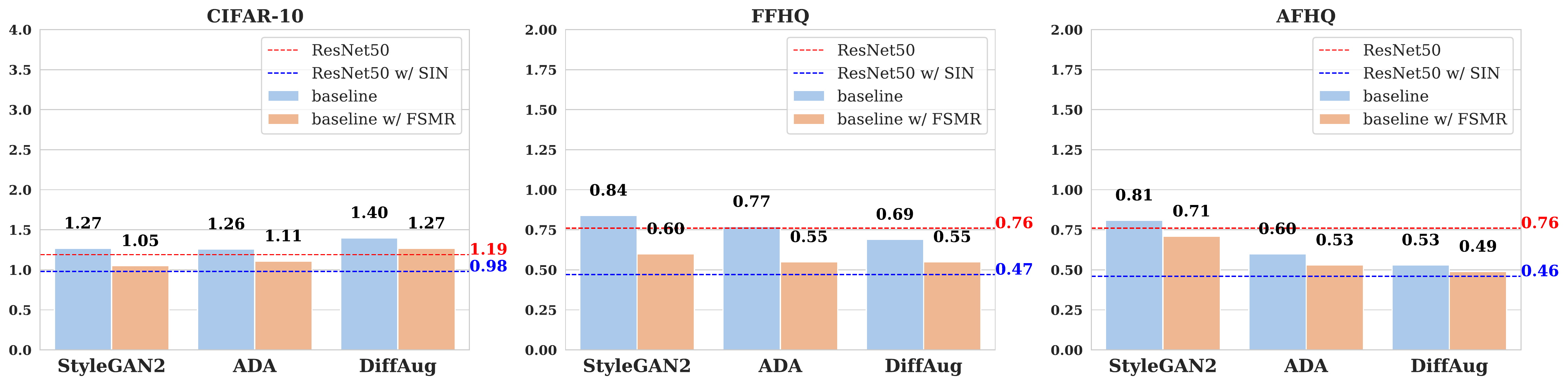} 
\caption{
\textbf{The relative distance of the discriminators}
on CIFAR-10, FFHQ, and AFHQ for StyleGAN2 variants. The higher $\rho$ value, the more sensitive the discriminator is to style when classifying real and fake. We report the reference values for the relative distances using ResNet50 trained on ImageNet (red line) and ResNet50 trained on Stylized ImageNet (blue line)\cite{geirhos2018imagenet}. As the lower relative distances agree with the higher classification performances, the lower relative distances of the discriminator agree with the higher generative performances.
}
\label{fig:fig5}
\end{figure*}
}

\newcommand{\figresultexample}{
\begin{figure*}[t]
\centering%
\newcommand{\hf}{0.16625\linewidth}%
\renewcommand{\h}{0.187\linewidth}%
\renewcommand{\hh}{\h*\real{2}/\real{3}}%
\renewcommand{\hhh}{\hf*\real{2}}%
\renewcommand{\hhhh}{\h*\real{2}*\real{6}/\real{20}}%
\renewcommand{\vv}{\vspace{-0.35mm}}%
\makebox[\hhh][c]{\textsc{FFHQ}}\hfill%
\makebox[\hh][c]{\textsc{MetFaces}}\hfill%
\makebox[\hh*\real{3}][c]{\textsc{AFHQ \FINAL{Cat}, Dog, \FINAL{Wild}, \scriptsize $256^2$}}\hfill%
\makebox[\hhhh][c]{\textsc{CIFAR-10}}\\%
\makebox[\hhh][c]{\scriptsize 70k img, $256^2$}\hfill%
\makebox[\hh][c]{\scriptsize 1336 img, $256^2$}\hfill%
\makebox[\hh][c]{\scriptsize \FINAL{5653 img}}%
\makebox[\hh][c]{\scriptsize 5239 img}%
\makebox[\hh][c]{\scriptsize \FINAL{5238 img}}\hfill%
\makebox[\hhhh][c]{\scriptsize 50k, 10 cls, $32^2$}\\%
\parbox[b]{\hf}{%
\includegraphics[width=\linewidth]{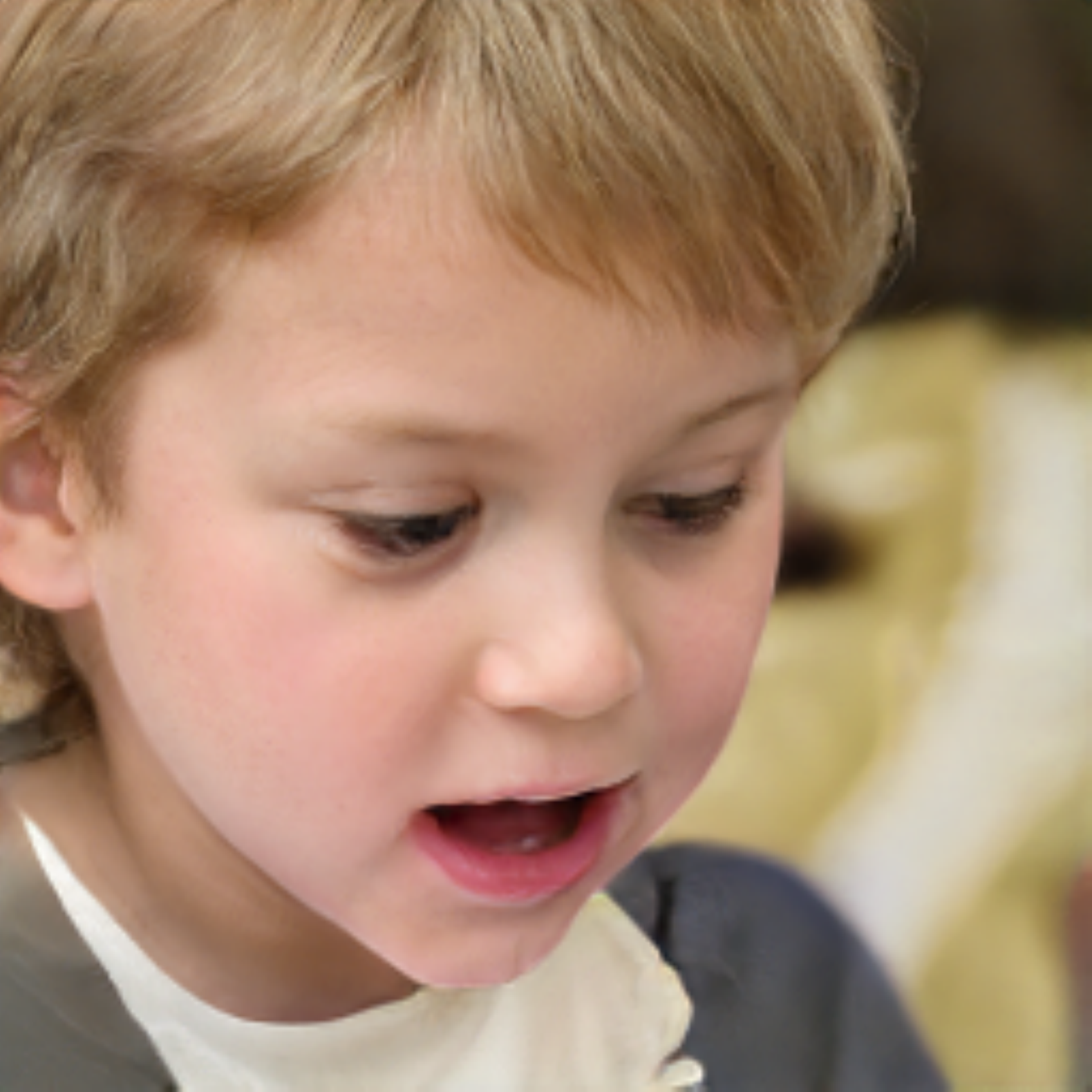}\vv\\
\includegraphics[width=\linewidth]{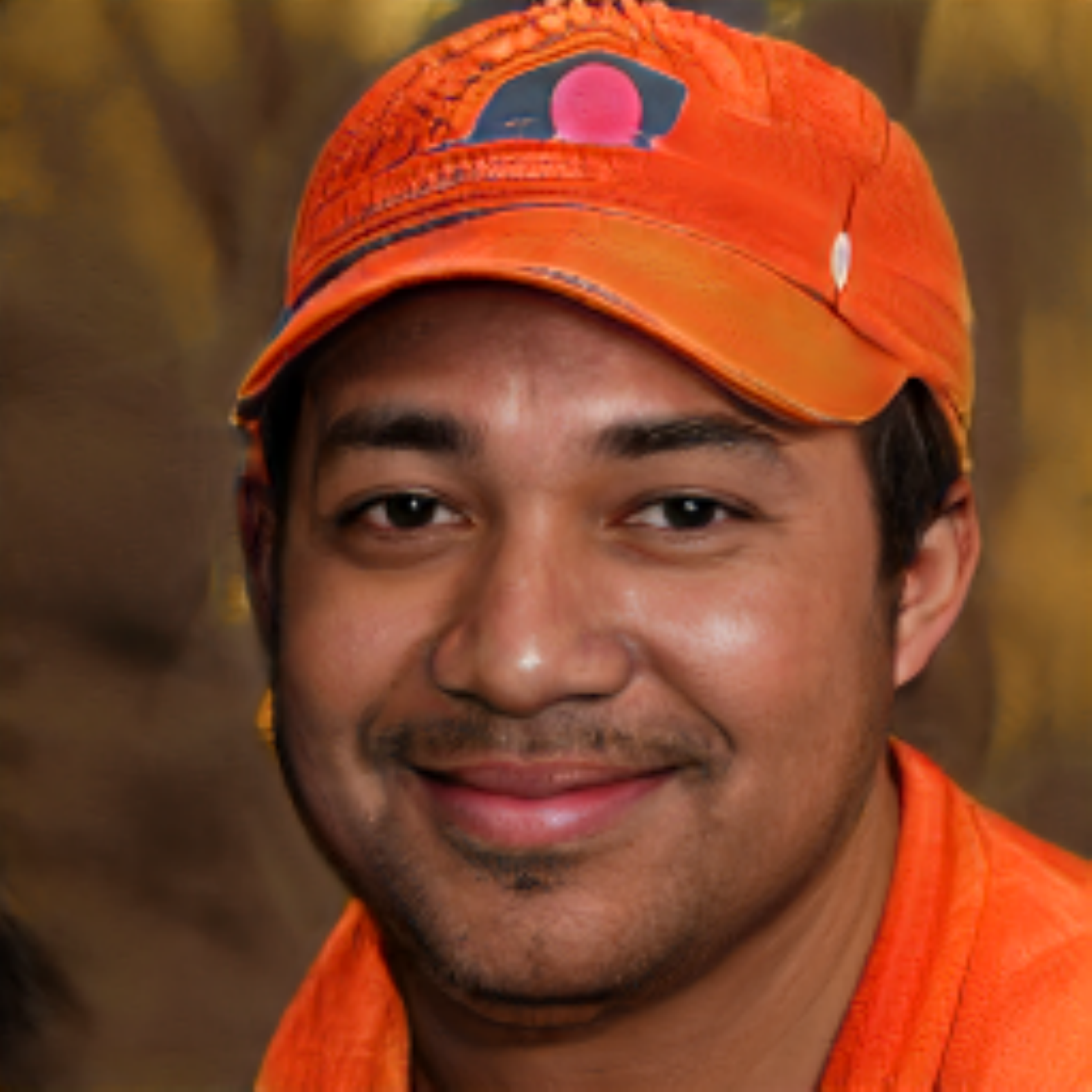}\vv\\
\includegraphics[width=\linewidth]{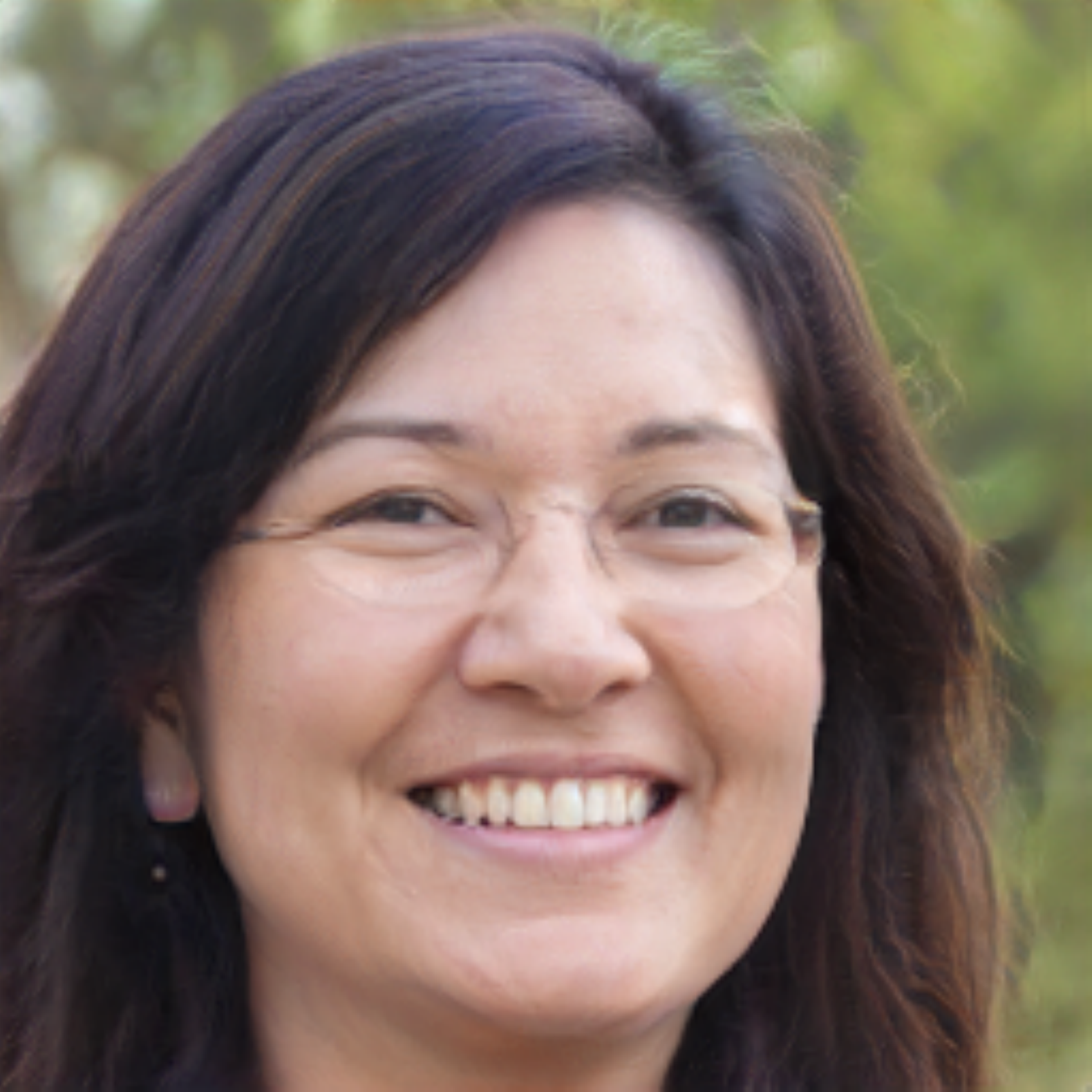}
}%
\parbox[b]{\hf}{%
\includegraphics[width=\linewidth]{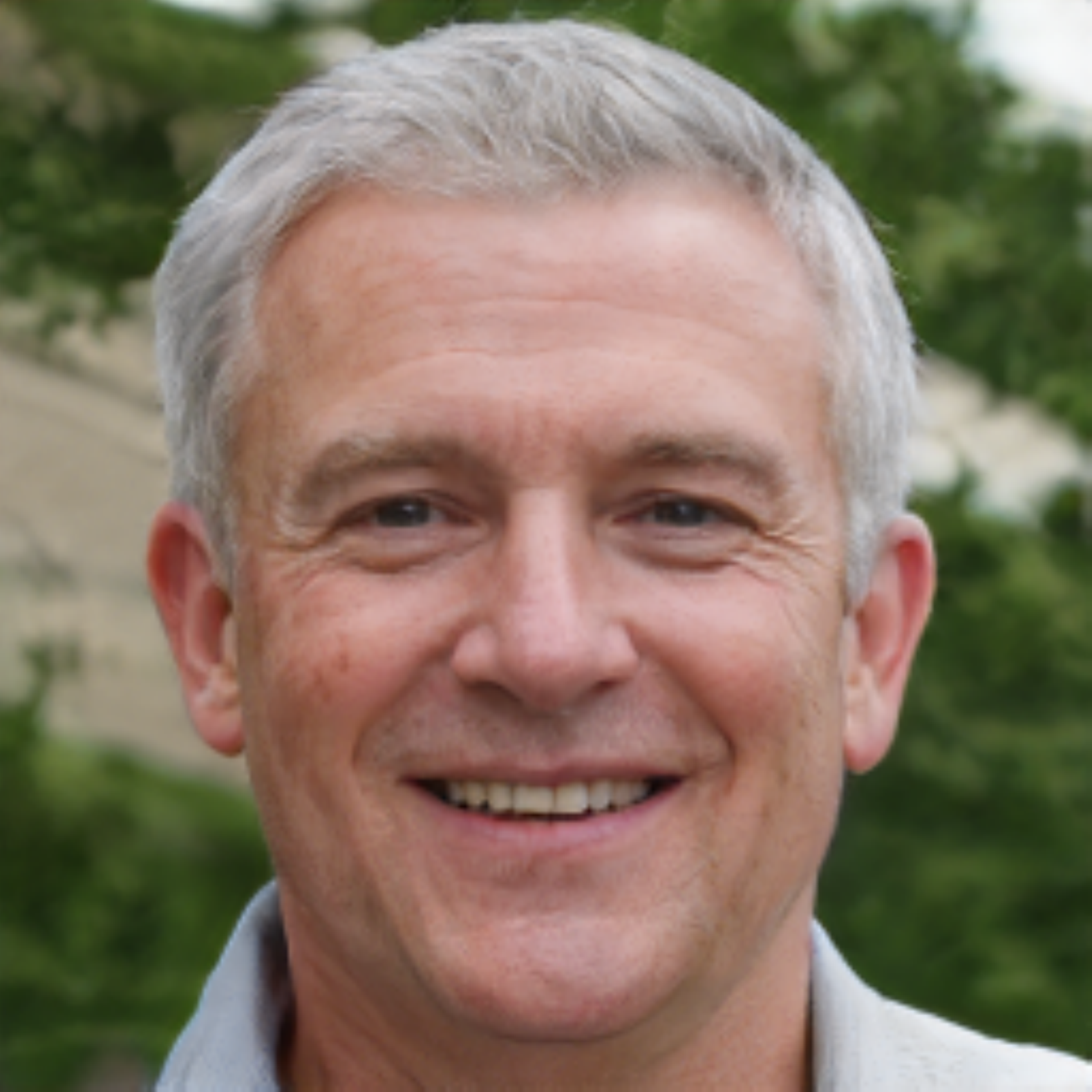}\vv\\
\includegraphics[width=\linewidth]{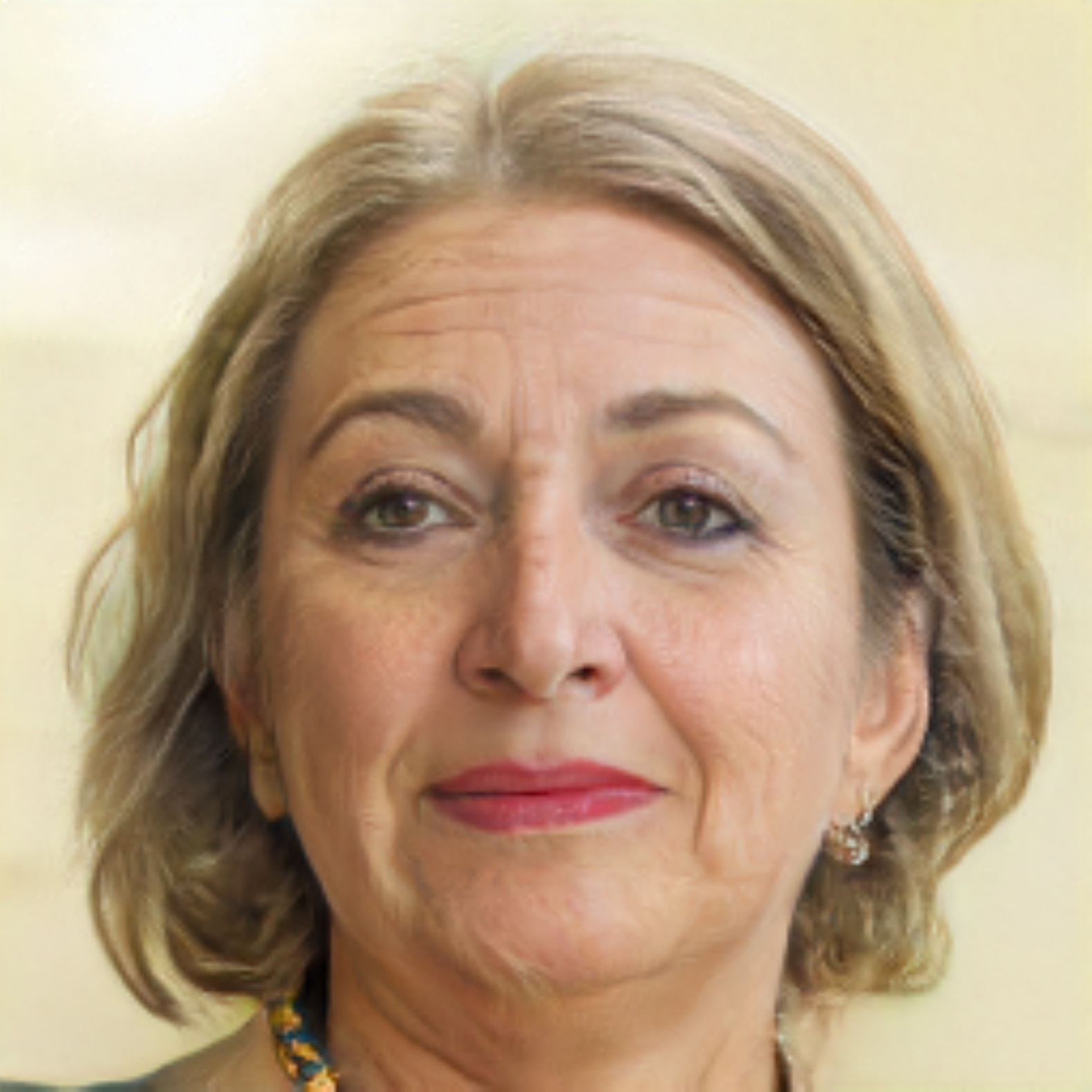}\vv\\
\includegraphics[width=\linewidth]{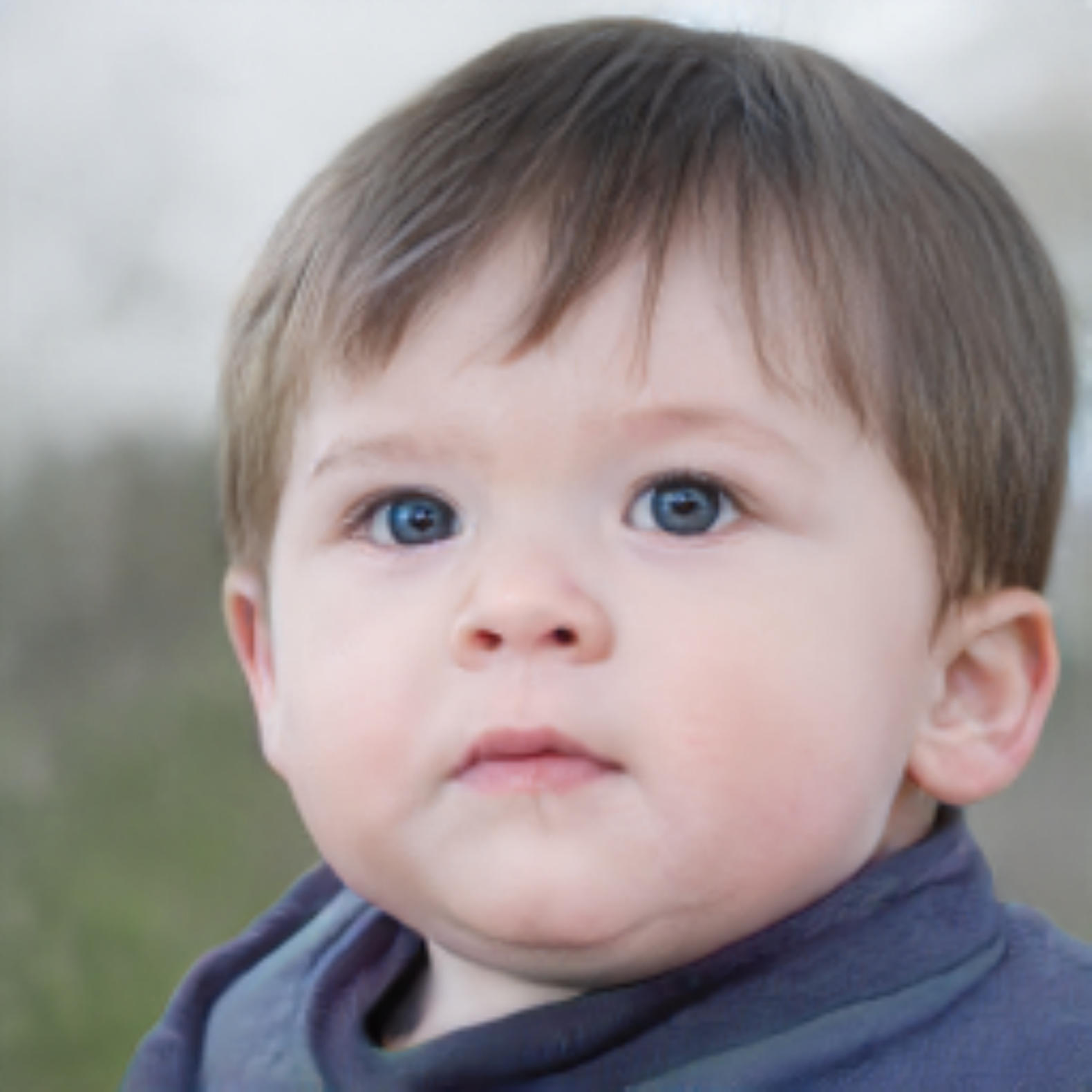}
}\hfill%
\parbox[b]{\hh}{%
\includegraphics[width=\linewidth]{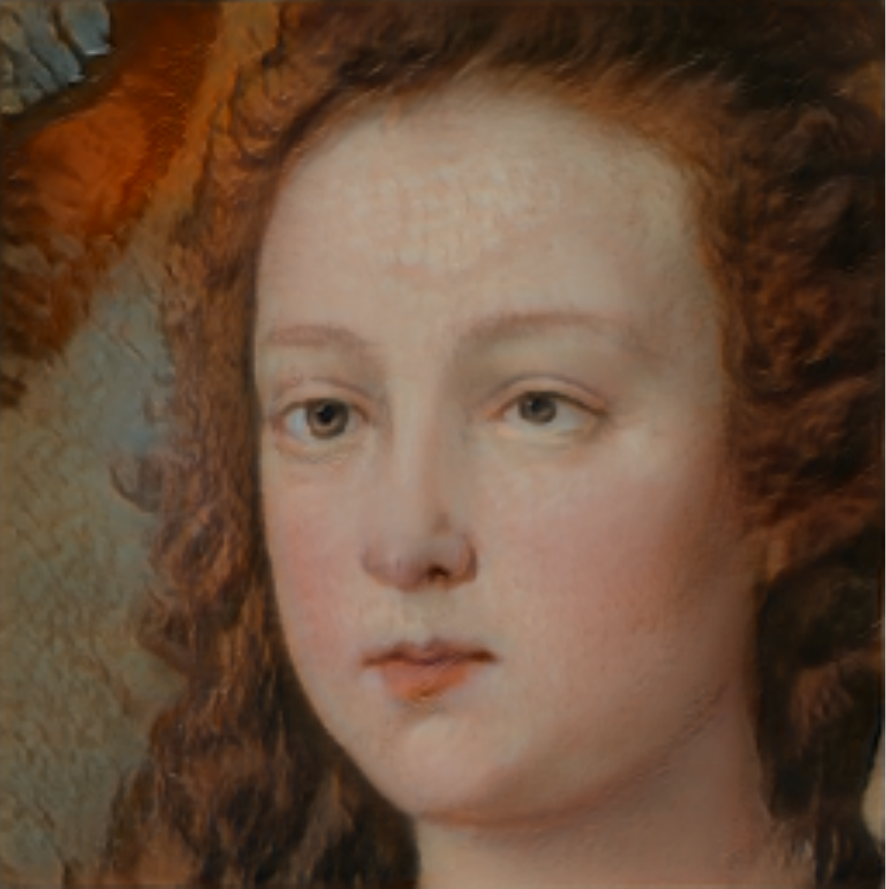}\vv\\
\includegraphics[width=\linewidth]{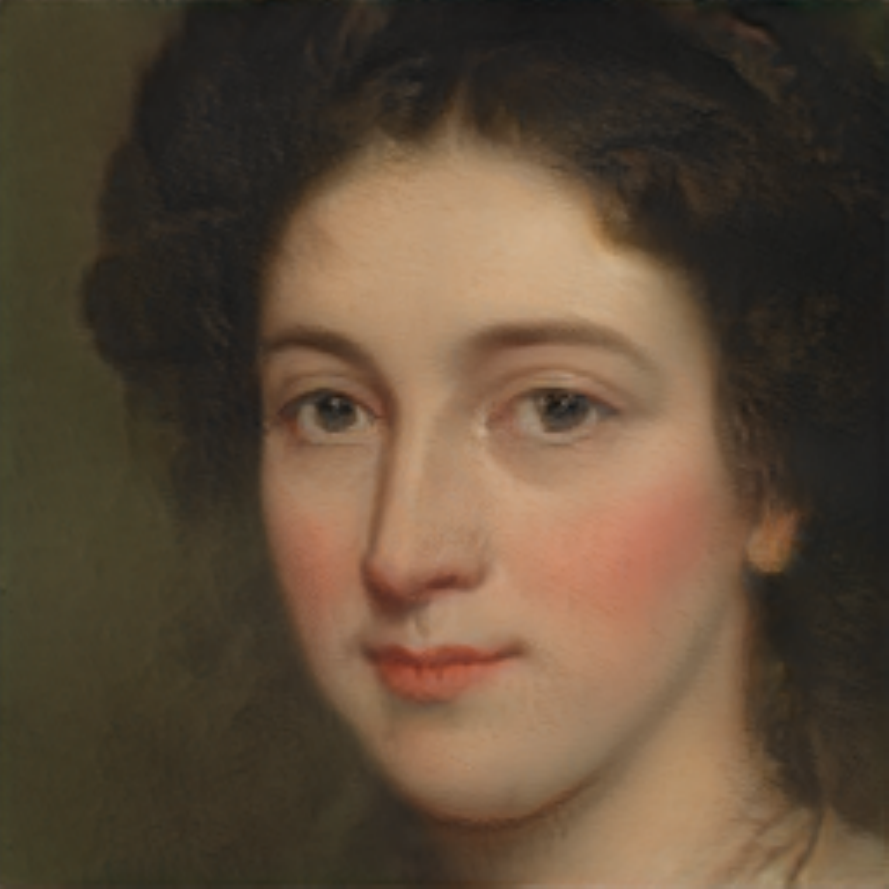}\vv\\
\includegraphics[width=\linewidth]{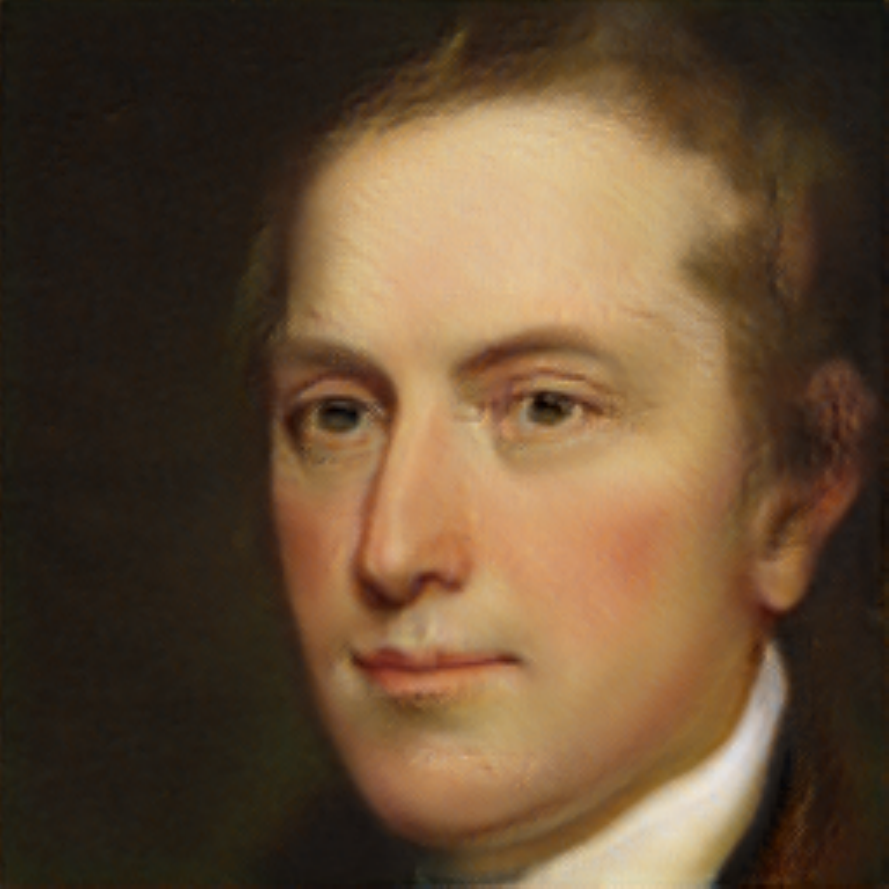}\vv\\
\includegraphics[width=\linewidth]{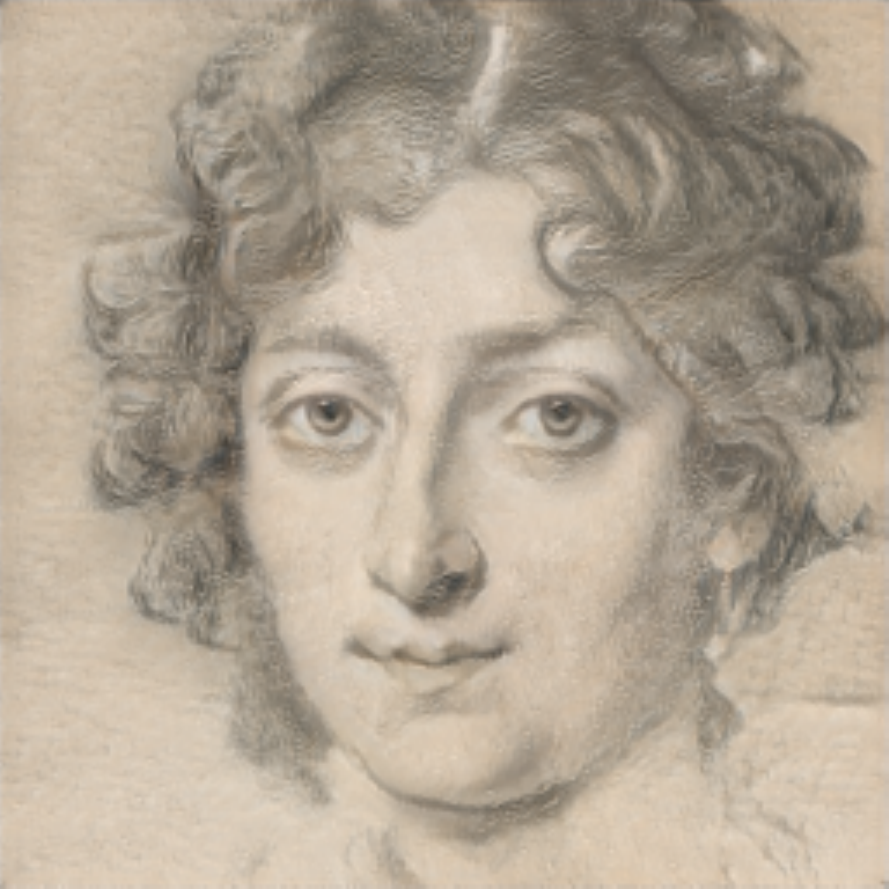}%
}\hfill%
\parbox[b]{\hh}{%
\includegraphics[width=\linewidth]{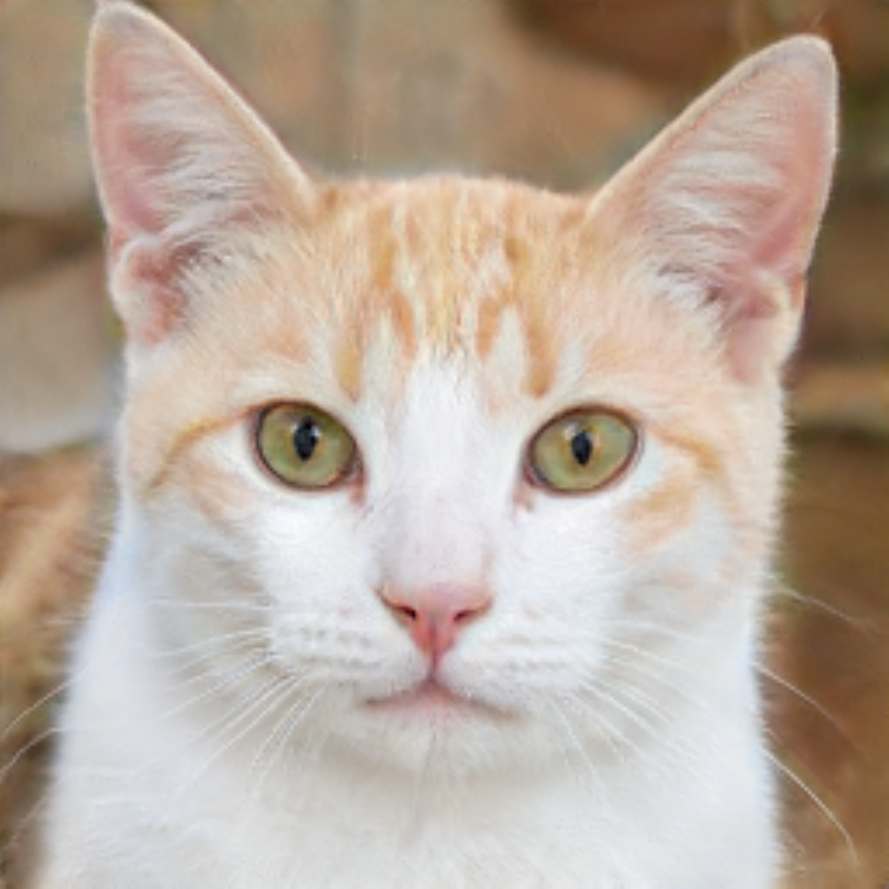}\vv\\
\includegraphics[width=\linewidth]{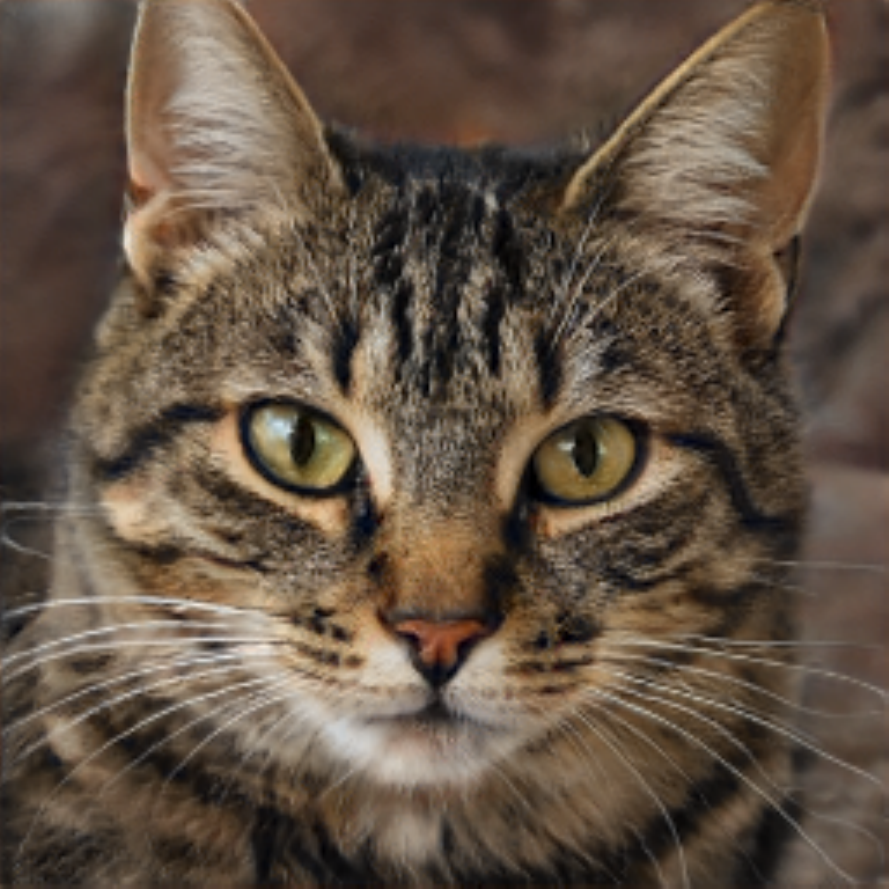}\vv\\
\includegraphics[width=\linewidth]{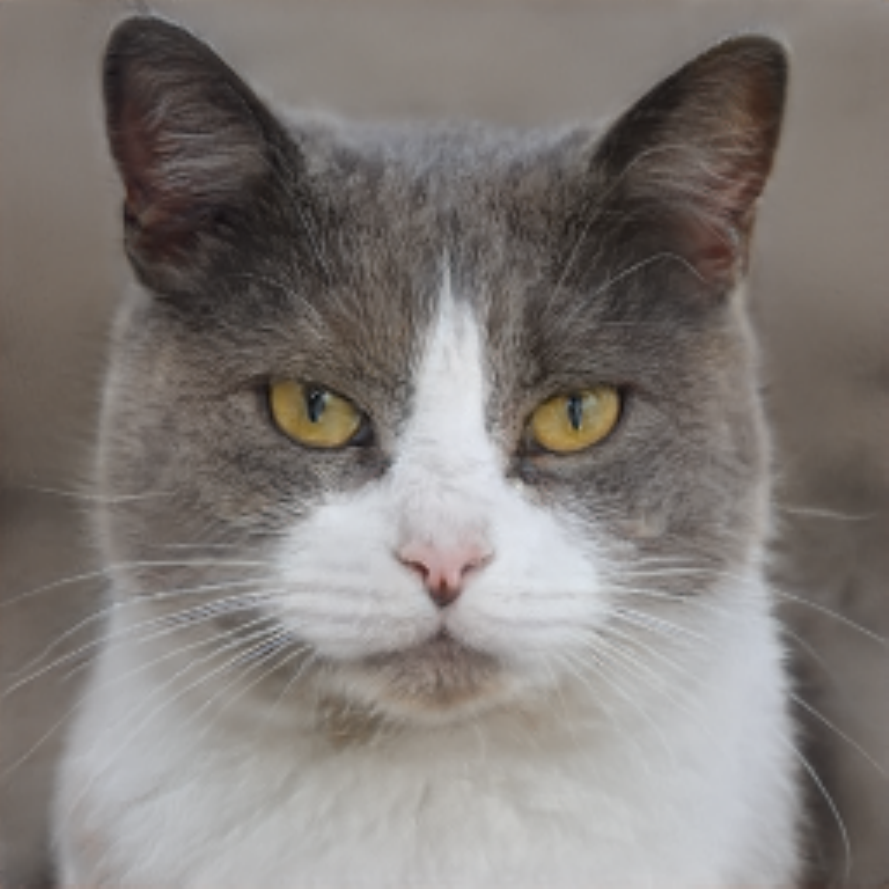}\vv\\
\includegraphics[width=\linewidth]{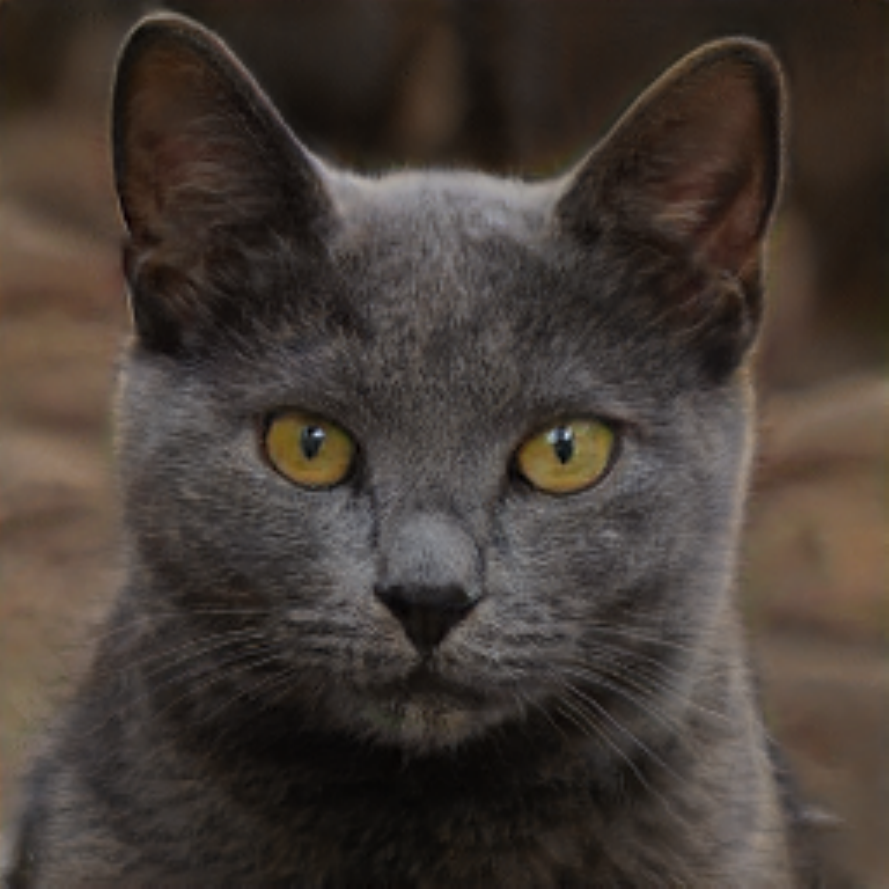}%
}%
\parbox[b]{\hh}{%
\includegraphics[width=\linewidth]{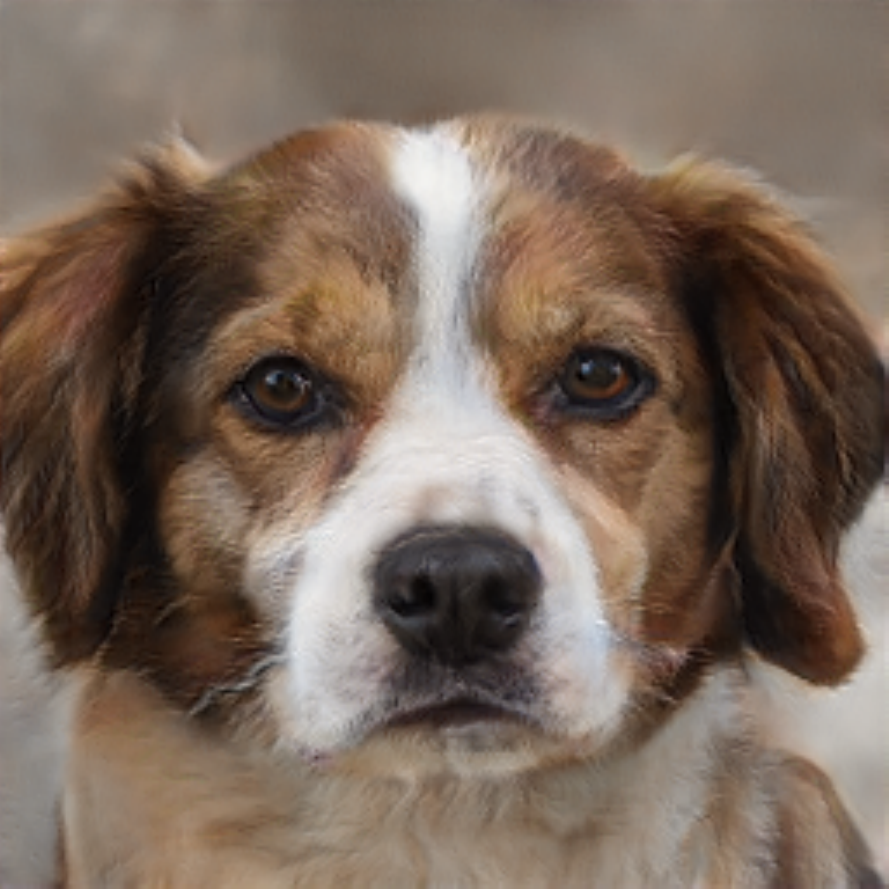}\vv\\
\includegraphics[width=\linewidth]{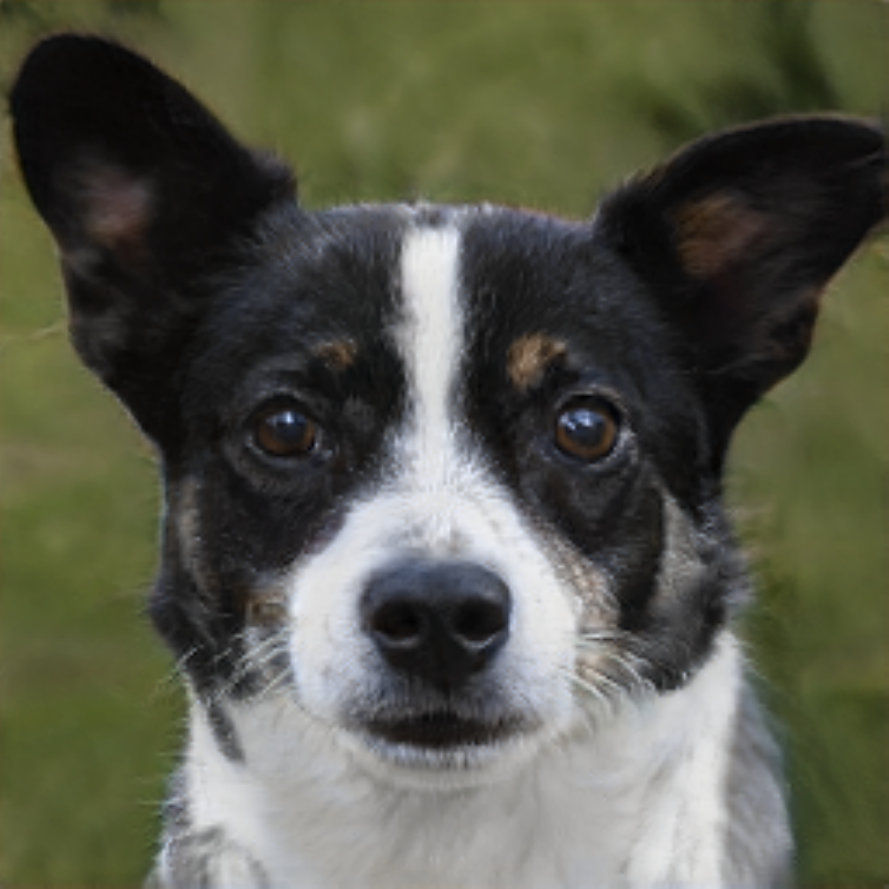}\vv\\
\includegraphics[width=\linewidth]{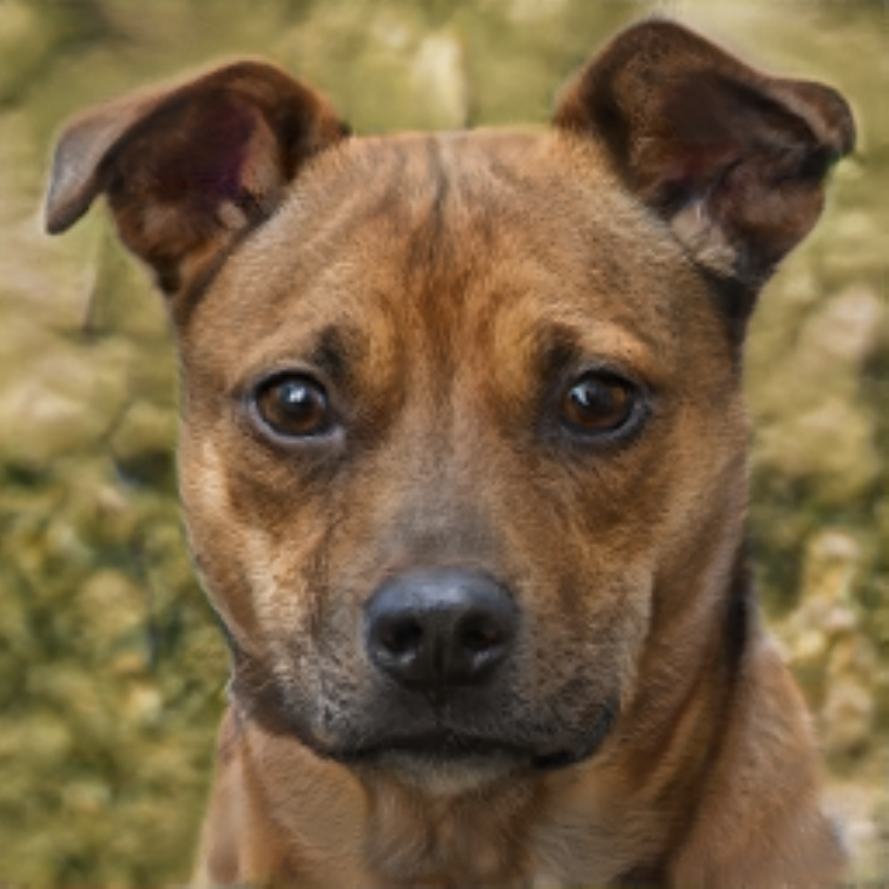}\vv\\
\includegraphics[width=\linewidth]{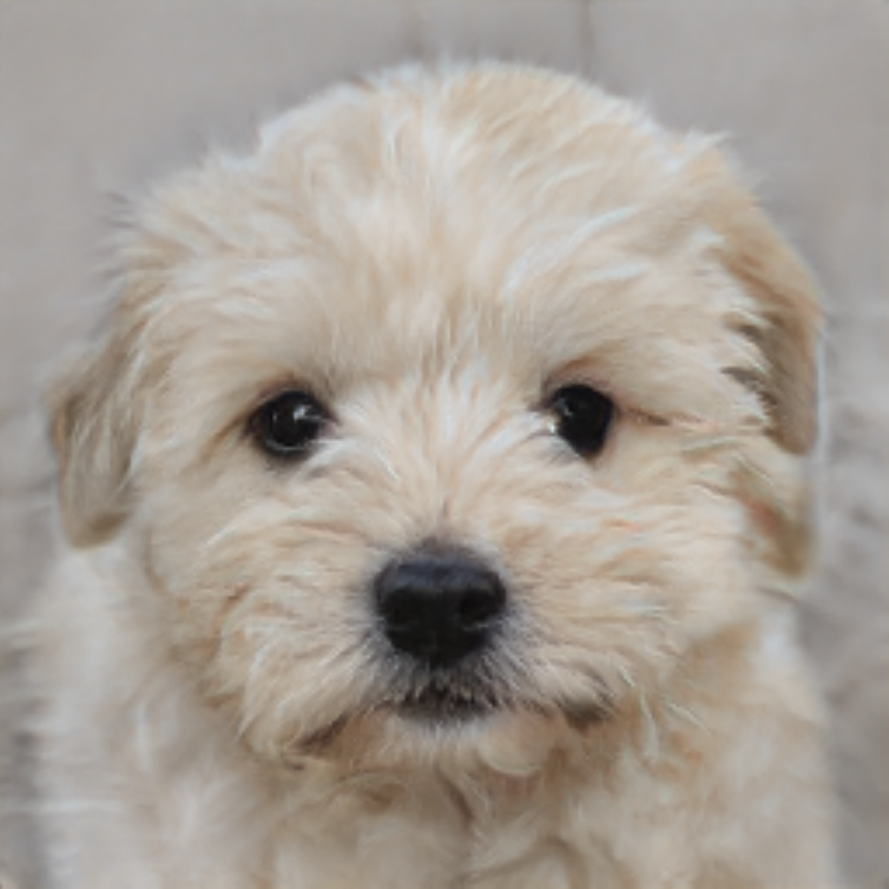}%
}%
\parbox[b]{\hh}{%
\includegraphics[width=\linewidth]{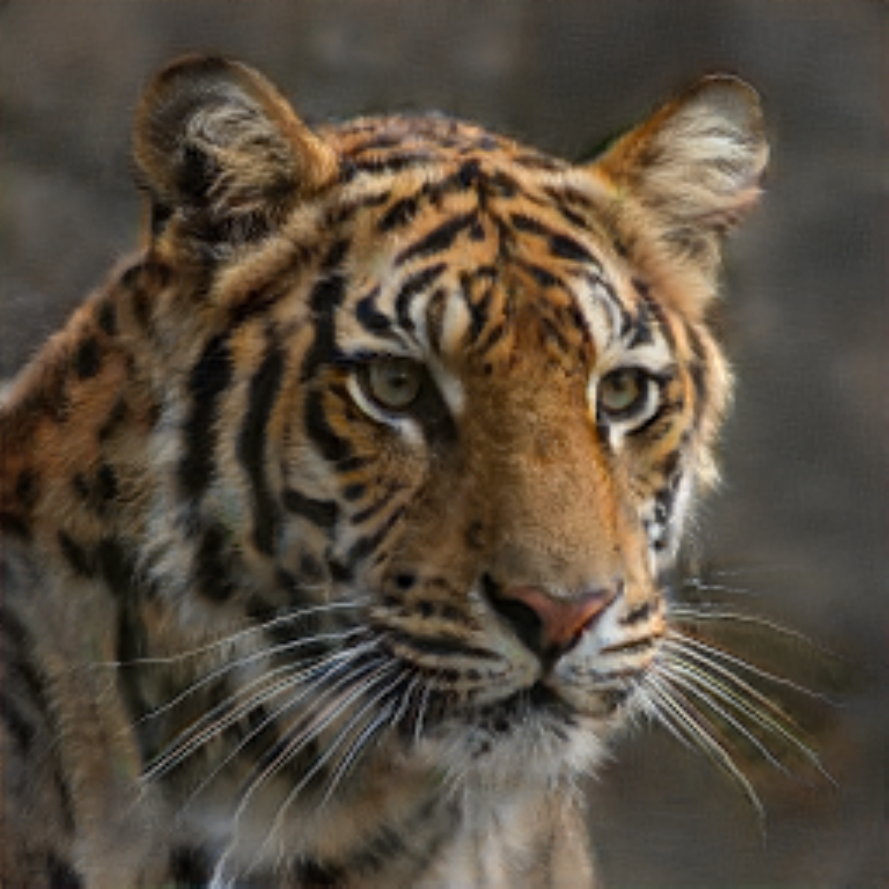}\vv\\
\includegraphics[width=\linewidth]{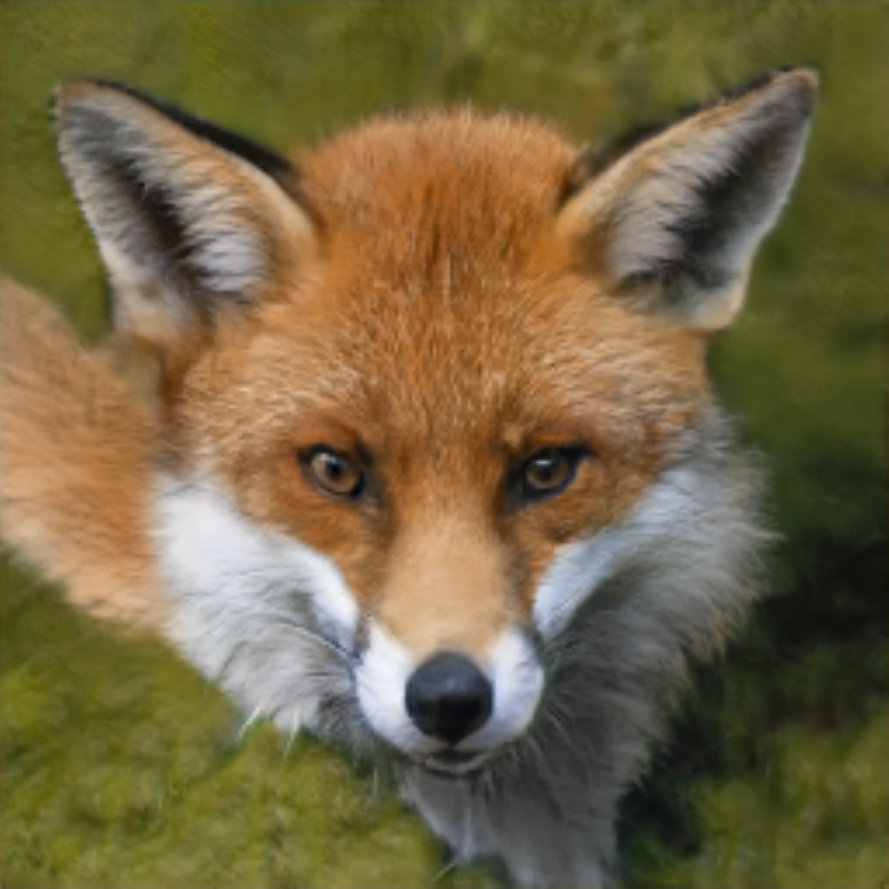}\vv\\
\includegraphics[width=\linewidth]{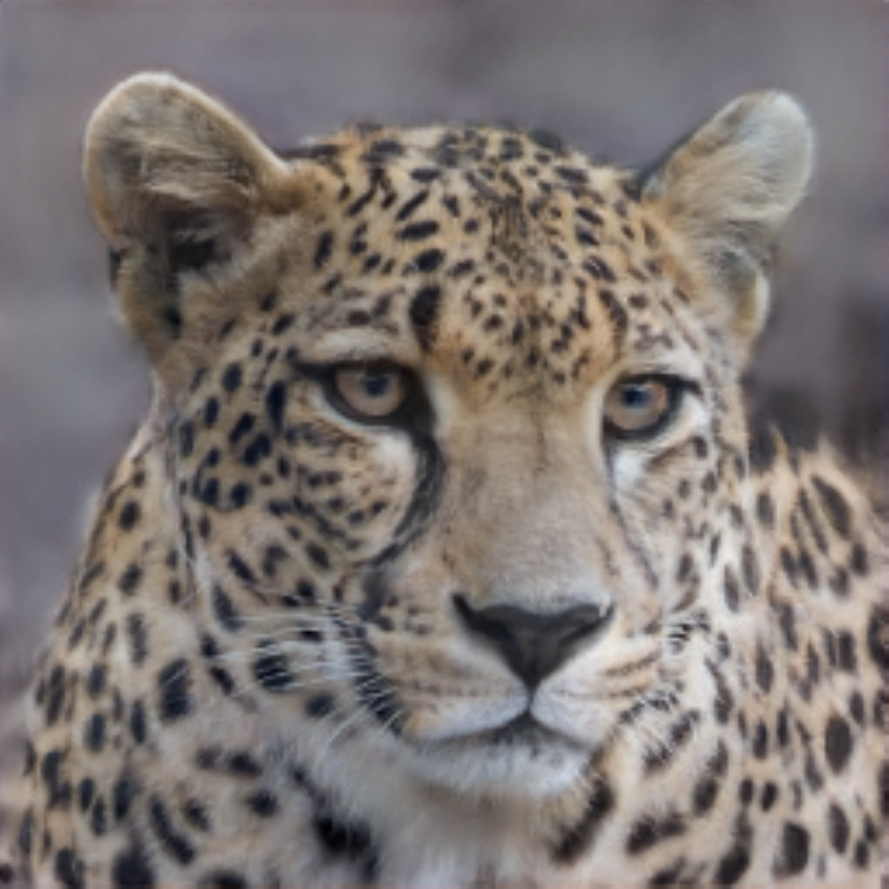}\vv\\
\includegraphics[width=\linewidth]{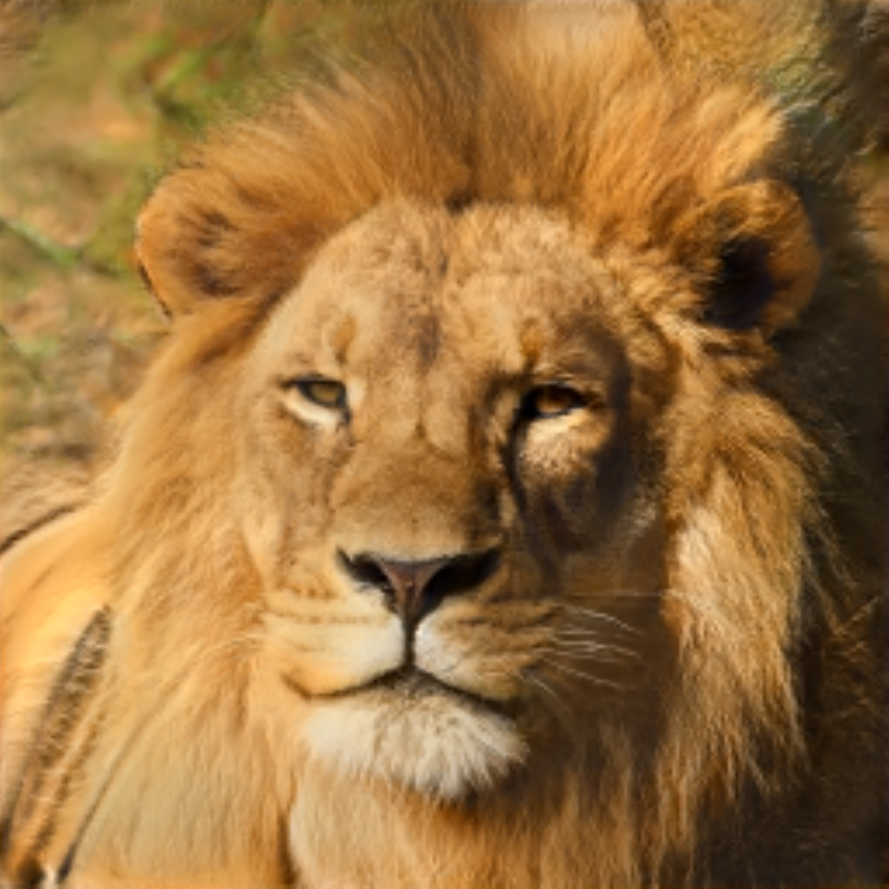}%
}\hfill%
\includegraphics[width=\hhhh]{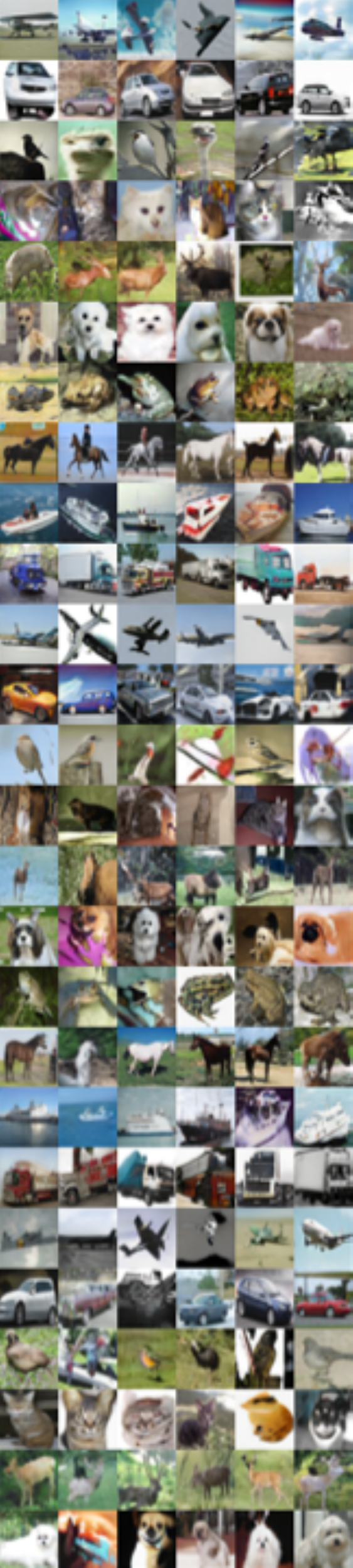}%
\caption{
Examples of generated images for several datasets trained using FSMR. Please note that we do not use transfer learning on MetFaces as opposed to ADA. See Appendix \ref{app:addresults} for more uncurated results.
}
\vspace{-3mm}
\label{fig:fig6}
\end{figure*}
}

\newcommand{\figsupgan}[1]{
\begin{figure*}[t]
\begin{center}
\includegraphics[width=1.0\linewidth]{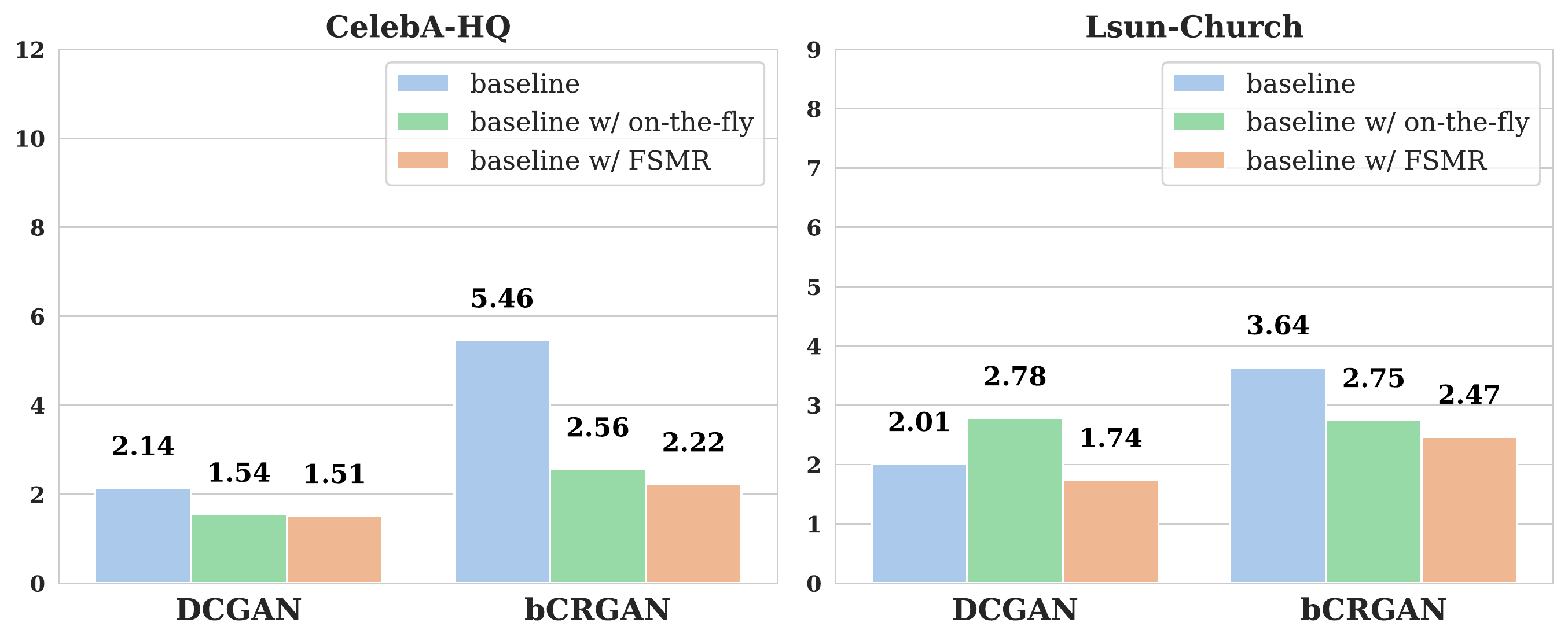}
\end{center}
\vspace{-6pt}
\caption{
\textbf{The relative distance of discriminator} results for several datasets.
}
\label{fig:figsupgan}
\end{figure*}
}

\newcommand{\figsupstyexam}[1]{
\begin{figure*}[t]
\begin{center}
\includegraphics[scale=0.25]{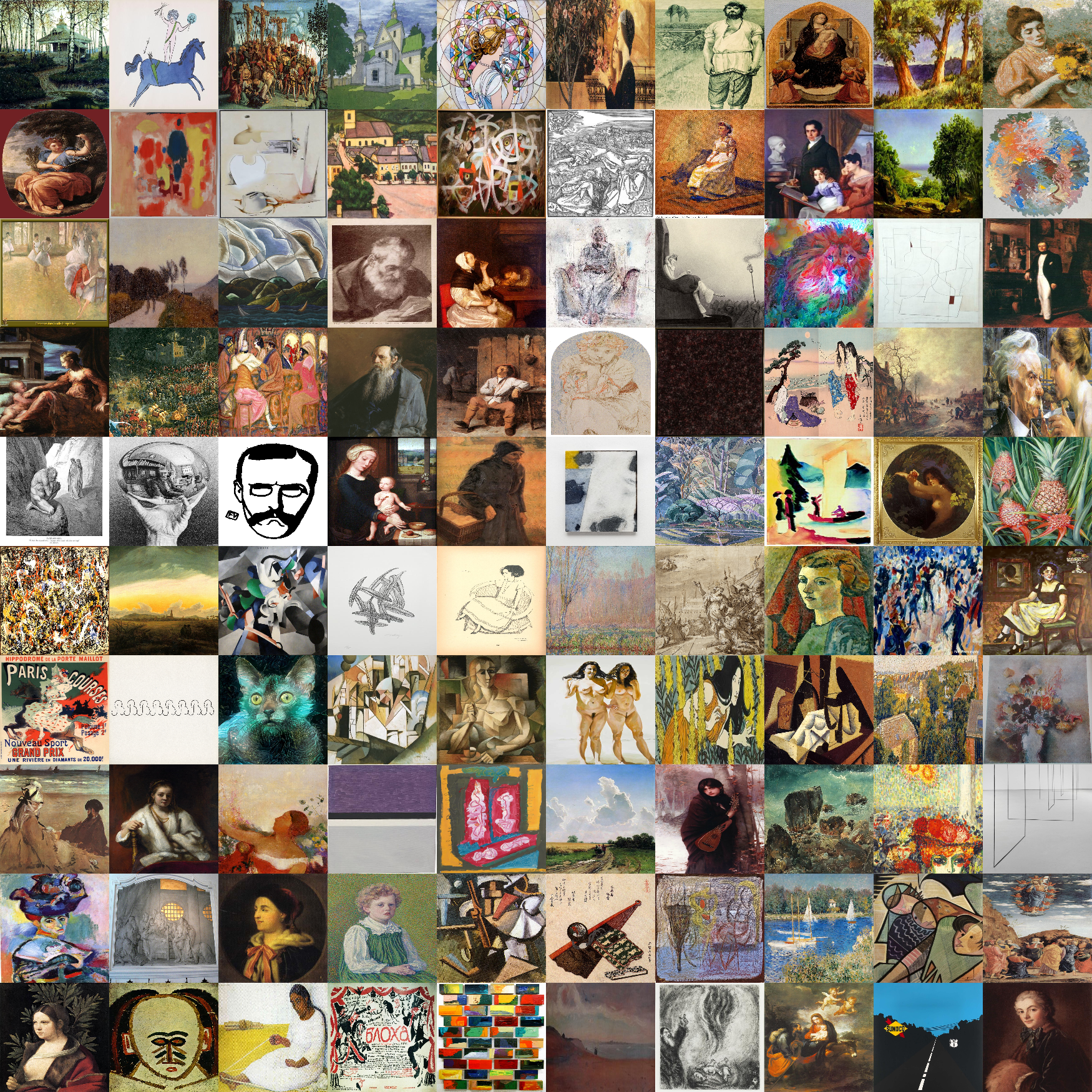}
\end{center}
\vspace{-6pt}
\caption{
\textbf{Example style images}. We present all examples of style images that we used in the comparison experiments.
}
\label{fig:figsupstyexam}
\end{figure*}
}

\newcommand{\figreconsup}[1]{
\begin{figure*}[t]
\centering
\newcommand{\ww}{0.08\linewidth}%
\newcommand{\w}{0.4\linewidth}%
\renewcommand{\h}{0.4\linewidth}%

\includegraphics[width=\ww,height=\h]{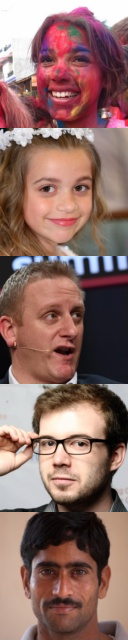}
\hspace{0.1mm}
\includegraphics[width=\w,height=\h]{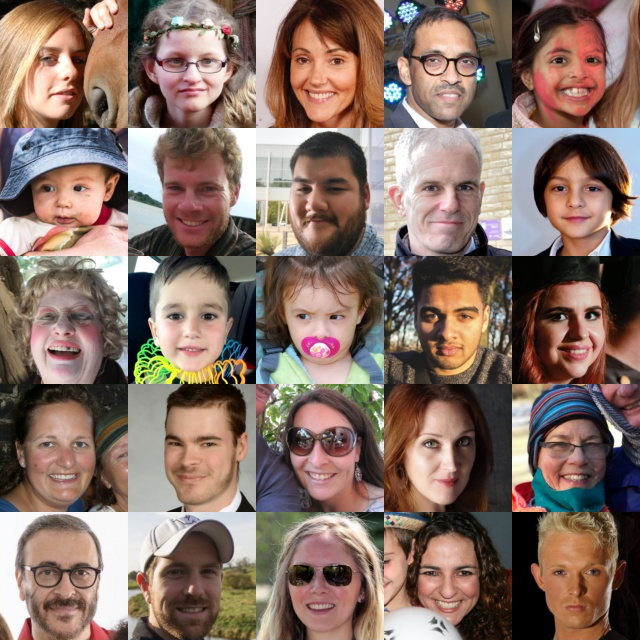}
\hspace{0.1mm}
\includegraphics[width=\w,height=\h]{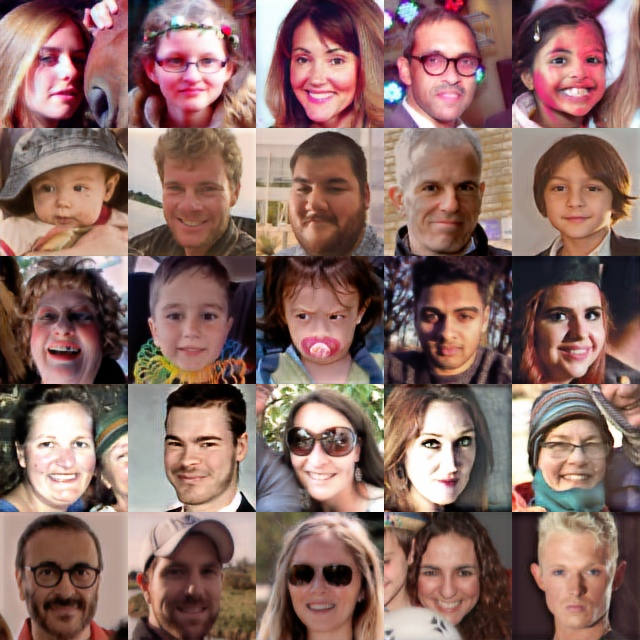}\vspace{1mm}\hfill%

\includegraphics[width=\ww,height=\h]{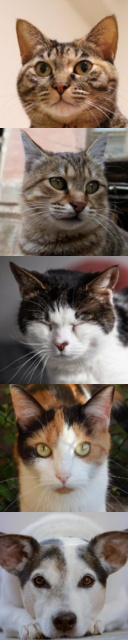}
\hspace{0.1mm}
\includegraphics[width=\w,height=\h]{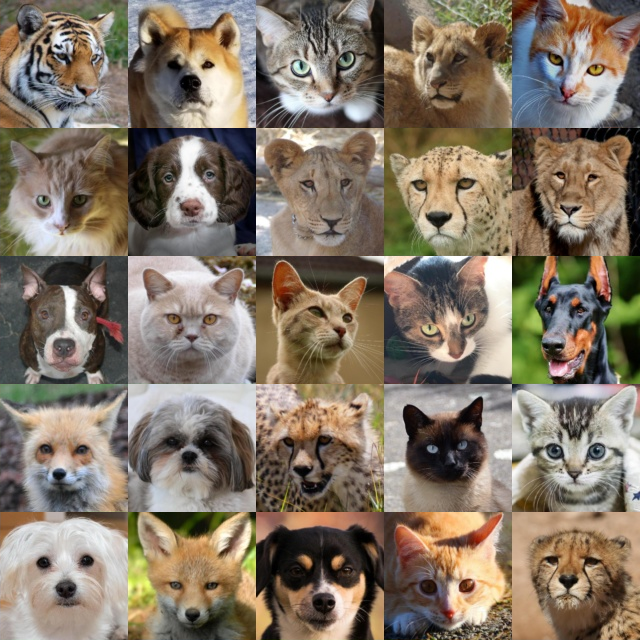}
\hspace{0.1mm}
\includegraphics[width=\w,height=\h]{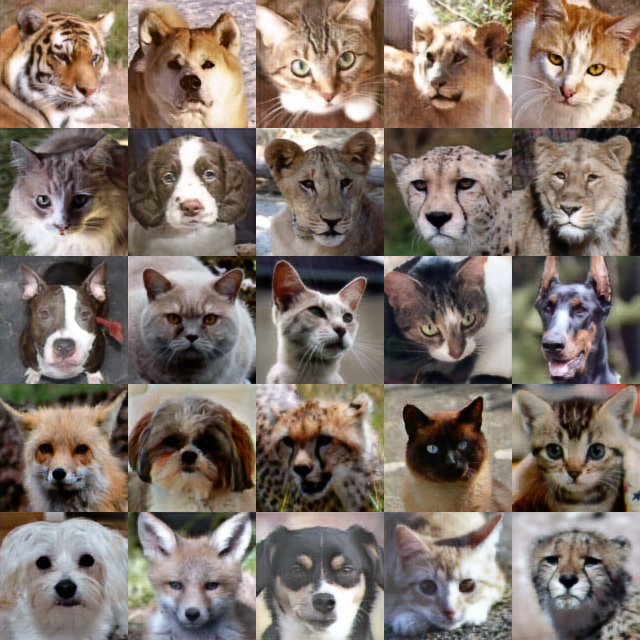}\vspace{1mm}\hfill%

\includegraphics[width=\ww,height=\h]{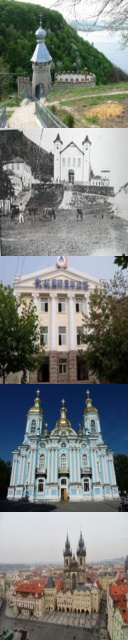}
\hspace{0.1mm}
\includegraphics[width=\w,height=\h]{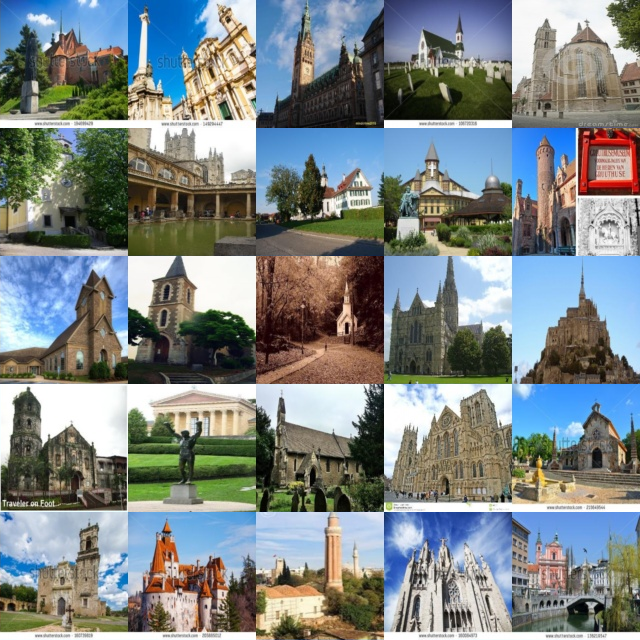}
\hspace{0.1mm}
\includegraphics[width=\w,height=\h]{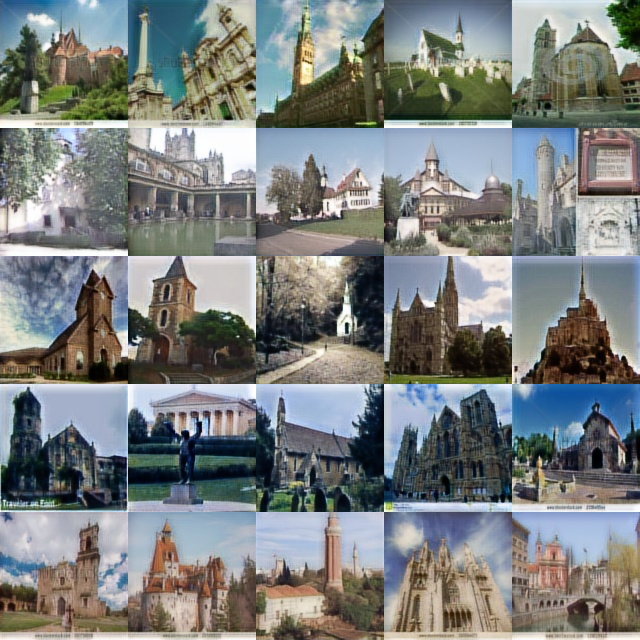}\hfill%

\makebox[\ww][c]{(a) Style}
\hspace{0.1mm}
\makebox[\w][c]{(b) Content}
\hspace{0.1mm}
\makebox[\w][c]{(d) Visualization of FSM}\hfill%

\caption{
\textbf{Additional visualization of the effect of FSM}. 
(a) Style images. (b) Content images. (c) Reconstruction of FSMed features preserves the detailed shapes. 
}
\label{fig:figreconsup}
\end{figure*}
}

\newcommand{\figsampleresult}[1]{
\begin{figure*}[t]
\centering
\includegraphics[scale=0.695]{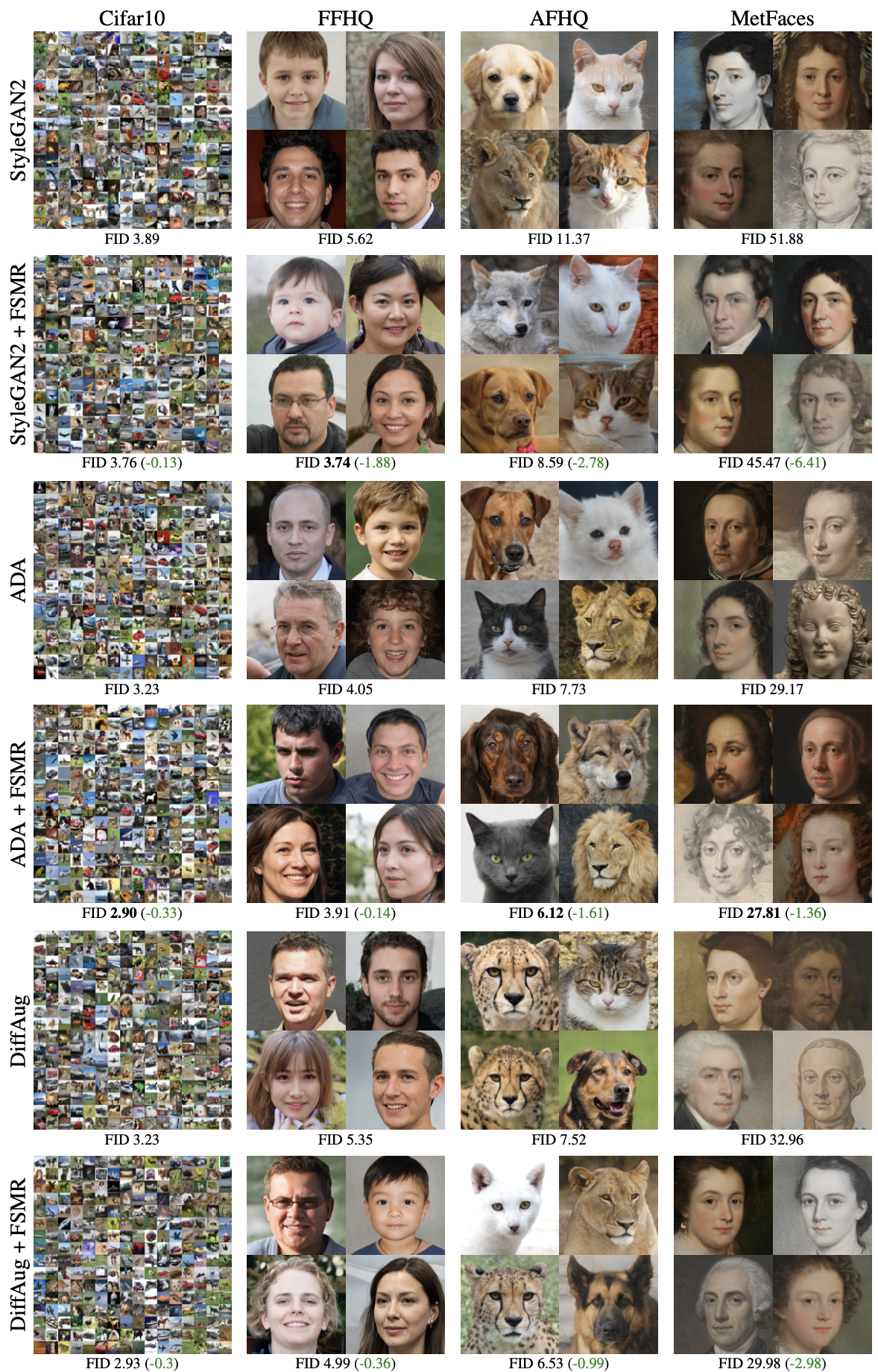} 
\caption{
Some example images and FIDs of competitors on CIFAR-10, FFHQ, AFHQ, and MetFaces.
}
\label{fig:figsampleresult}
\end{figure*}
}

\newcommand{\figuncurateall}{
\begin{figure*}[t]
\centering
\vspace{-7mm}
\makebox[0.32\linewidth][c]{\textbf{FFHQ}, FID: \textbf{3.76}}
\makebox[0.32\linewidth][c]{\textbf{MetFaces}, FID: \textbf{45.47}}
\makebox[0.32\linewidth][c]{\textbf{AFHQ}, FID: \textbf{8.59}}\\
\makebox[0mm][l]{\rotatebox{90}{\makebox[0.4\linewidth][c]{\textbf{StyleGAN2 + FSMR}}}}\hspace{2.8mm}
\hspace{0.1mm}\includegraphics[width=0.32\linewidth]{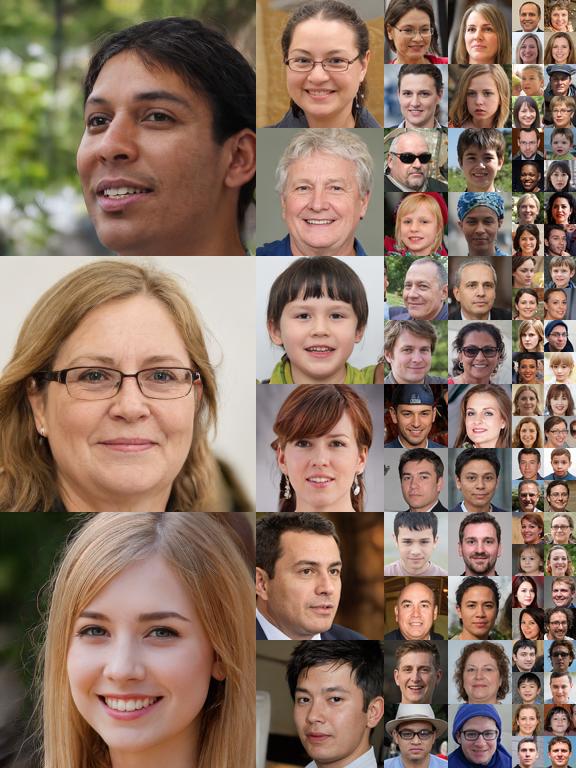}
\hspace{0.1mm}\includegraphics[width=0.32\linewidth]{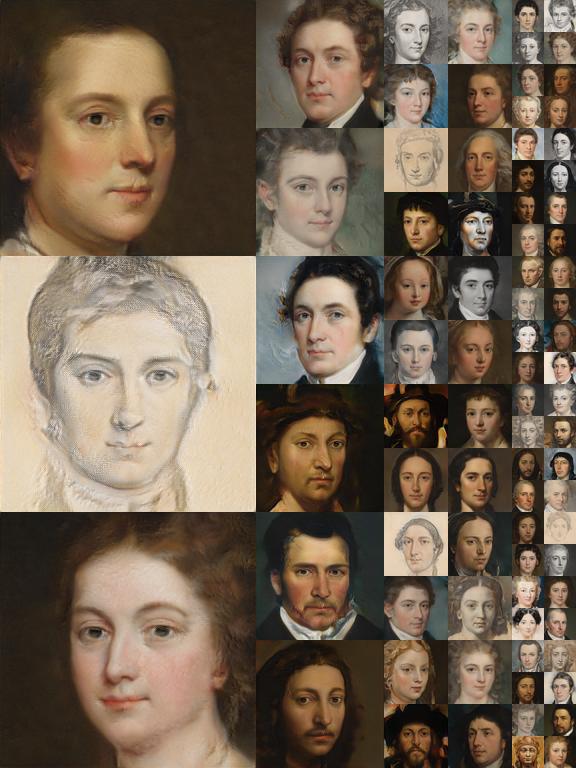}
\hspace{0.1mm}\includegraphics[width=0.32\linewidth]{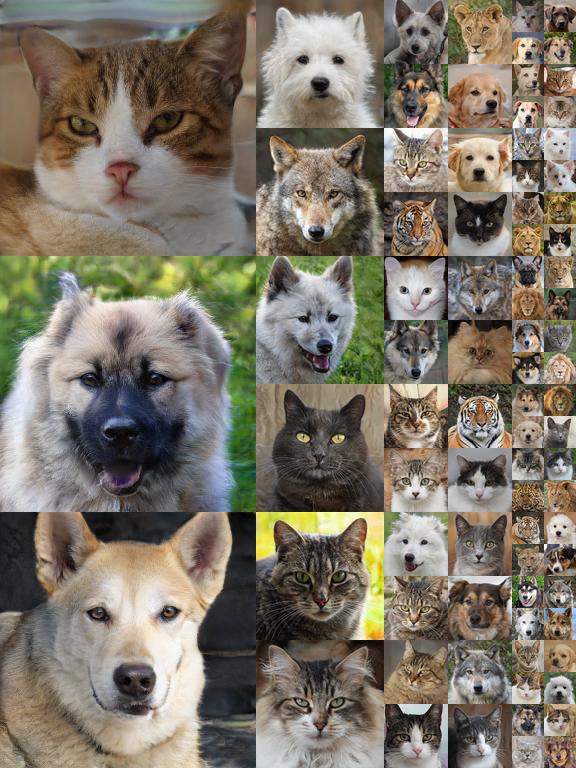}\\

\makebox[0.32\linewidth][c]{\textbf{FFHQ}, FID: \textbf{3.91}}
\makebox[0.32\linewidth][c]{\textbf{MetFaces}, FID: \textbf{27.81}} 
\makebox[0.32\linewidth][c]{\textbf{AFHQ}, FID: \textbf{6.12}}\\
\makebox[0mm][l]{\rotatebox{90}{\makebox[0.4\linewidth][c]{\textbf{ADA + FSMR}}}}\hspace{2.8mm}
\hspace{0.1mm}\includegraphics[width=0.32\linewidth]{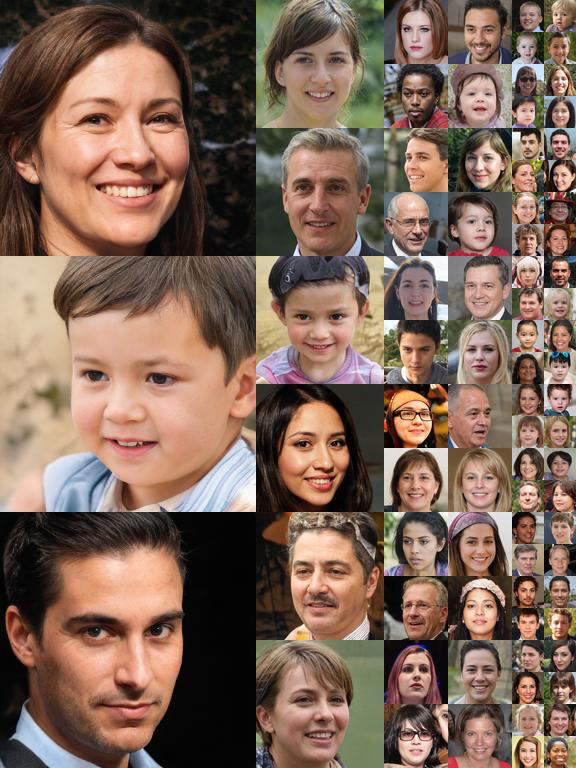}
\hspace{0.1mm}\includegraphics[width=0.32\linewidth]{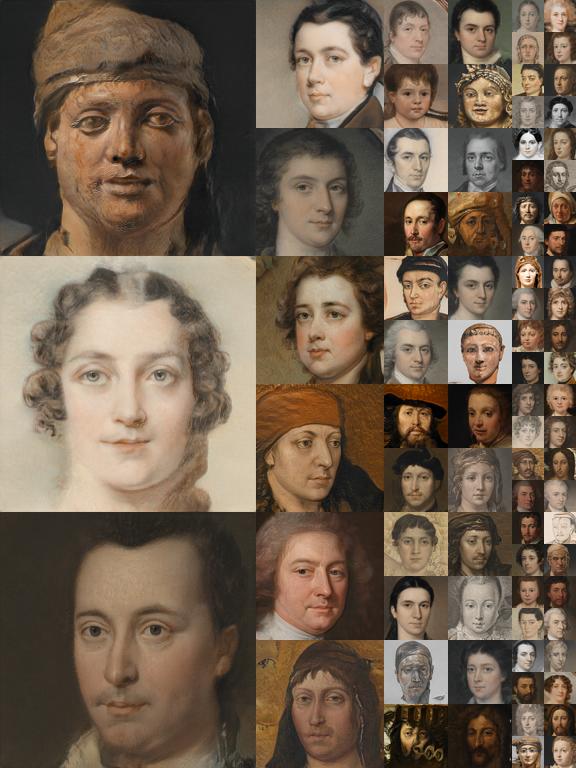}
\hspace{0.1mm}\includegraphics[width=0.32\linewidth]{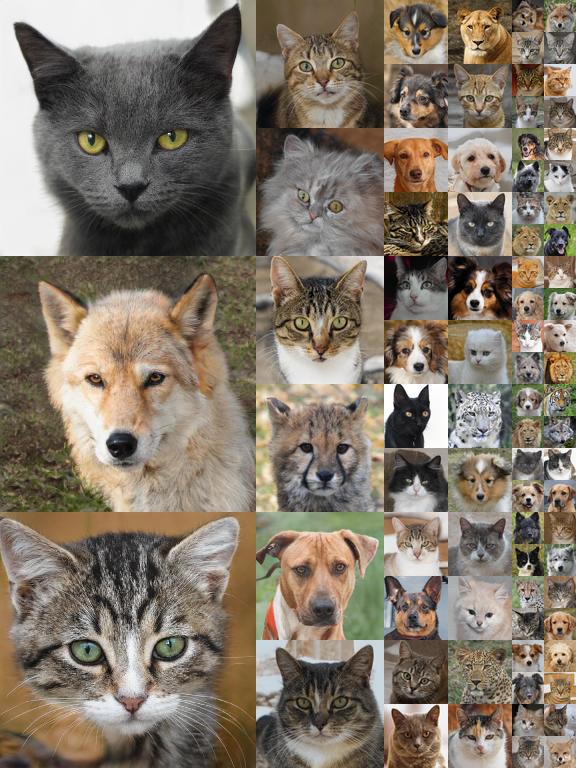}\\

\makebox[0.32\linewidth][c]{\textbf{FFHQ}, FID: \textbf{4.99}}
\makebox[0.32\linewidth][c]{\textbf{MetFaces}, FID: \textbf{29.98}}
\makebox[0.32\linewidth][c]{\textbf{AFHQ}, FID: \textbf{6.53}}\\
\makebox[0mm][l]{\rotatebox{90}{\makebox[0.4\linewidth][c]{\textbf{DiffAug + FSMR}}}}\hspace{2.8mm}
\hspace{0.1mm}\includegraphics[width=0.32\linewidth]{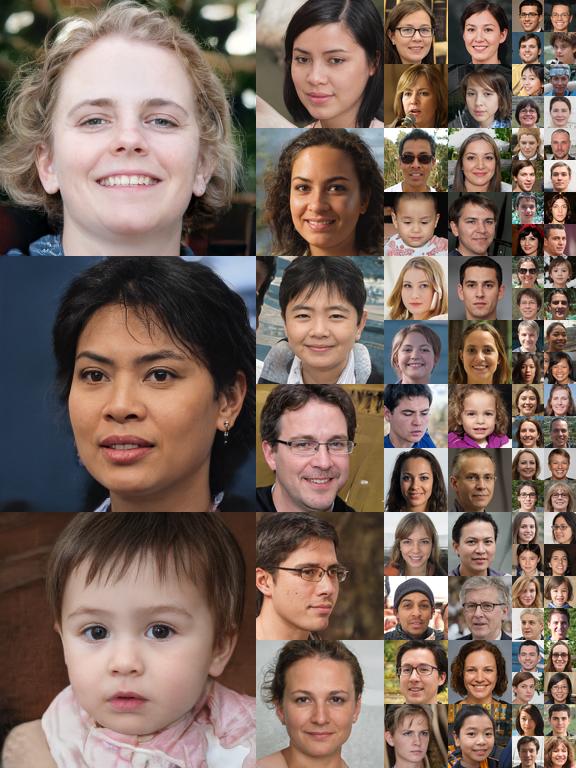}
\hspace{0.1mm}\includegraphics[width=0.32\linewidth]{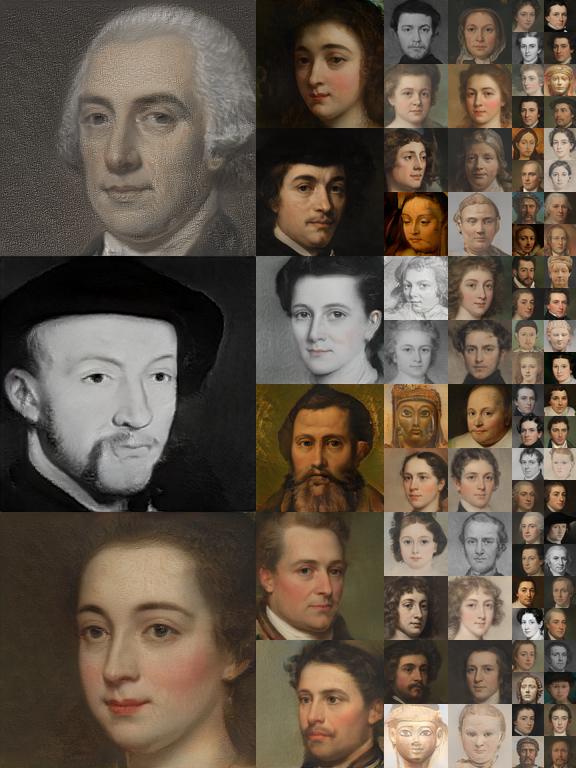}
\hspace{0.1mm}\includegraphics[width=0.32\linewidth]{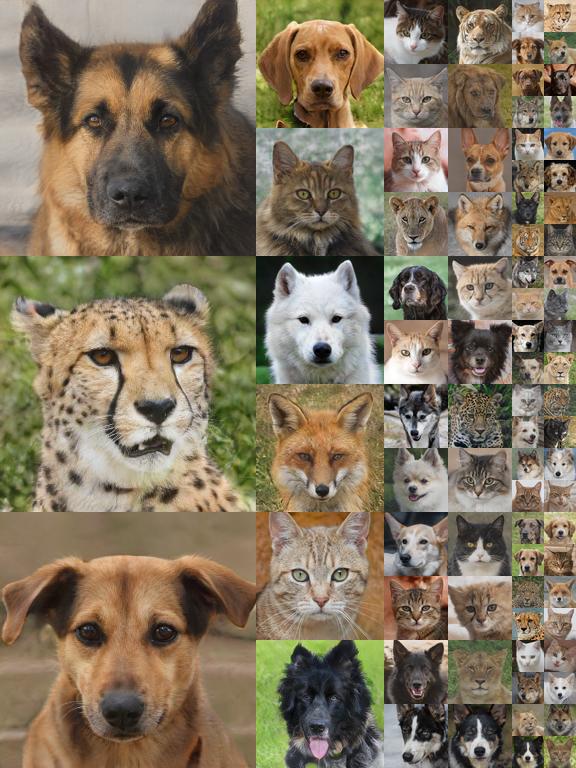}\\

\vspace{3mm}
\caption{Uncurated random samples from the models trained with FSMR in three datasets.} 
\label{fig:figuncurateall}
\end{figure*}
}
\newcommand{\tabfidgan}{
\begin{table*}[t]
\begin{center}
\small{
\begin{tabular}{l c c c c c c c}
\toprule
\multirow{2}{*}{Method} & \multicolumn{5}{c}{Standard dataset} & \multicolumn{2}{c}{Costs}\\
\cmidrule(lr){2-6}
\cmidrule(lr){7-8}
{} & {CIFAR-10} & {FFHQ} & {AFHQ} & {CelebA-HQ} & {LSUN Church} & {Time \footnotesize{(Hours)}} & {Memory \footnotesize{(GB)}} \\
\midrule
DCGAN   &15.89\std{0.12}    &7.82\std{0.10}   &17.27\std{0.13} &6.71\std{0.09} &17.33\std{0.11} &25.4 (1.5$^\dagger$) &5 (4$^\dagger$)\\
DCGAN w/ on-the-fly   &15.88\std{0.11}	& 7.33\std{0.17}	&14.22\std{0.15} &5.41\std{0.10}	&26.05\std{0.14} &31.5 (1.8$^\dagger$) &8.5 (7.5$^\dagger$)\\
DCGAN w/ FSMR   &\B 14.98\std{0.09} 	&\B 6.76\std{0.08}	&\B 13.19\std{0.09}	&\B 5.23\std{0.10} &\B 13.84\std{0.10} &26.2 (1.6$^\dagger$) &5.1 (4$^\dagger$)\\
\midrule
bCRGAN  &12.46\std{0.09}   &6.43\std{0.08}   &9.35\std{0.10} &4.31\std{0.09}  &13.20\std{0.10} &26.1 (1.6$^\dagger$) &5 (4$^\dagger$)\\
bCRGAN w/ on-the-fly   &12.43\std{0.10}  	&5.20\std{0.09}	&8.63\std{0.12} &3.47\std{0.09}	&10.51\std{0.10} &33.1 (1.9$^\dagger$) &8.5 (7.5$^\dagger$)\\
bCRGAN w/ FSMR   &\B 11.17\std{0.07}    &\B 4.68\std{0.08}	&\B 8.33\std{0.08}  &\B 3.43\std{0.09}	&\B 9.09\std{0.07} &27.7 (1.7$^\dagger$) &5.1 (4$^\dagger$)\\
\bottomrule
\end{tabular}
}
\end{center}
\caption{
\textbf{FID comparison on DCGAN variants} with FSMR and the baseline on-the-fly stylization.
The bold numbers indicate the best FID for each baseline. We report the mean FID over 3 training runs together with standard deviations and the additional costs. All image resolutions are set to $128 \times 128$ due to the backbone architecture except CIFAR-10 ($32\times32$). Time and memory are measured in $128\times128$ images, and $\dagger$ indicates what is measured in $32\times32$ images. Time means a full training time.
}

\label{tab:table1}
\end{table*}
}

\newcommand{\tabfidstylegan}{
\begin{table*}[t]
\setlength{\tabcolsep}{4pt}
\begin{center}
\begin{tabular}{l c c c c c c c}
\toprule
\multirow{2}{*}{Method} & \multicolumn{3}{c}{Standard dataset} & \multicolumn{4}{c}{Small dataset} \\
\cmidrule(lr){2-4}
\cmidrule(lr){5-8}
{} & {CIFAR-10} & {FFHQ} & {AFHQ} & {MetFaces} & {AFHQ Dog} & {AFHQ Cat} & {AFHQ Wild} \\
\midrule
StyleGAN2 &3.89\std{0.07} &5.62\std{0.10} &11.37\std{0.03}	&51.88\std{0.44}	&19.65\std{0.07}	&8.37\std{0.06} &4.17\std{0.06}  \\
+ \methodshort &\B 3.76\std{0.03}	&\B 3.74\std{0.03}	&\B 8.59\std{0.03}	&\B 45.47\std{0.42}	&\B 18.08\std{0.07}	&\B 6.69\std{0.04}  &\B 3.96\std{0.03}  \\
\midrule
StyleGAN2-ADA &3.23\std{0.06} & 4.05\std{0.07} &7.73\std{0.11}	&29.17\std{0.08}	&13.56\std{0.10}	&6.64\std{0.09} &3.74\std{0.14}  \\
+ \methodshort &\B 2.90\std{0.08}	&\B 3.91\std{0.06}	&\B 6.12\std{0.10}	&\B 27.81\std{0.11}	&\B 11.76\std{0.14}	&\B 5.71\std{0.10}    &\B 3.24\std{0.16}  \\
\midrule
StyleGAN2-DiffAug &3.23\std{0.08} &5.35\std{0.09} &7.52\std{0.08}	&32.96\std{0.08}	&16.92\std{0.06}	&6.39\std{0.05} &4.39\std{0.07}  \\
+ \methodshort &\B 2.93\std{0.05}	&\B 4.99\std{0.08}	&\B 6.53\std{0.05}	&\B 29.98\std{0.15}	&\B 14.55\std{0.18}	&\B 6.29\std{0.07} &\B 4.28\std{0.04}  \\
\bottomrule
\end{tabular}
\end{center}
\caption{
\textbf{FID comparison on StyleGAN2 variants}.
 The bold numbers indicate the best FID for each baseline. We report the mean FID over 3 training runs together with standard deviations. FSMR improves the baselines in all cases.
}
\label{tab:table2}
\end{table*}
}

\newcommand{\tabCIFAR}{%
\begin{table}[t]
\begin{center}
\small{
\begin{tabular}{l c c c c}
\toprule
\multirow{2}{*}{Method} & \multicolumn{2}{c}{Unconditional} & \multicolumn{2}{c}{Conditional} \\
\cmidrule(lr){2-3}
\cmidrule(lr){4-5}
& FID $\downarrow$ & \s IS $\uparrow$ & FID $\downarrow$ & \s IS $\uparrow$ \\
\midrule
\FINAL{StyleGAN2}         & \FINAL{\s3.89}       & \FINAL{\s9.36}       & \FINAL{\s3.52}       & \FINAL{\s9.77}       \\
\FINAL{+ \methodshort}         & \FINAL{\s\textbf{3.76}}       & \FINAL{\s\textbf{9.58}}       & \FINAL{\s\textbf{3.35}}       & \FINAL{\s\textbf{10.05}}       \\
\midrule
\FINAL{StyleGAN2-ADA}         & \FINAL{\s3.23}       & \FINAL{\s9.47}       & \FINAL{\s2.76}       & \FINAL{\s9.98}       \\
\FINAL{+ \methodshort}         & \FINAL{\s\textbf{2.90}}       & \FINAL{\s\textbf{9.68}}       & \FINAL{\s\textbf{2.63}}       & \FINAL{\s\textbf{10.03}}       \\
\midrule
\FINAL{StyleGAN2-DiffAug}         & \FINAL{\s3.23}       & \FINAL{\s9.63}       & \FINAL{\s3.10}       & \FINAL{\s9.84}       \\
\FINAL{+ \methodshort}         & \FINAL{\s\textbf{2.93}}       & \FINAL{\s\textbf{9.81}}       & \FINAL{\s\textbf{2.87}}       & \FINAL{\s\textbf{10.02}}       \\
\bottomrule
\end{tabular}
}
\end{center}
\caption{\textbf{FID and inception score comparison} on CIFAR-10 across StyleGAN2 variants. Bold face indicates the best scores for each baseline. We report the mean scores over three training runs.}
\label{tab:table3}
\end{table}
}

\newcommand{\tabstytime}{
\begin{table}[t]
\begin{center}
\scriptsize{
\begin{tabular}{l c c c c c}

\toprule
{} & {CIFAR-10} & {CelebA-HQ} & {FFHQ} & {AFHQ} & {LSUN Church} \\
\midrule
Time   &8  &10   &30   &5  &40 \\
\bottomrule
\end{tabular}
}
\end{center}
\caption{
\textbf{The time to create the stylized dataset} for each standard dataset, measured in hours.
}
\label{tab:tabstytime}
\end{table}
}

\newcommand{\retabmix}{
\begin{table}[h]
\begin{center}
\small{
\begin{tabular}{l c c c}
{} & {CIFAR-10} & {FFHQ} & {AFHQ} \\
\midrule
ADA w/ FSMR   &\textbf{2.90}  &\textbf{3.91}   &\textbf{6.12} \\
\midrule
ADA w/ Mixup &3.48 &4.40 &6.67 \\
\midrule
ADA w/ CutMix &3.45 &4.36 &6.51 \\
\bottomrule
\end{tabular}
}
\end{center}
\vspace{-4mm}
\caption{\textbf{Comparison from the previous mixing methods.}}
\label{tab:retab1}
\end{table}
}

\newcommand{\retabfly}{
\begin{table}[h]
\begin{center}
\scriptsize{
\begin{tabular}{l c c c}
{} & {CIFAR-10} & {FFHQ} & {AFHQ} \\
\midrule
DCGAN   &16.02 / 15.88 / \textbf{14.98} &7.52 / 7.33 / \textbf{6.76}   &14.87 / 14.22 / \textbf{13.19} \\
\midrule
bCRGAN &12.58 / 12.43 / \textbf{11.17} &5.74 / 5.20 / \textbf{4.68} &9.02 / 8.63 / \textbf{8.33} \\
\bottomrule
\end{tabular}
}
\end{center}
\vspace{-4mm}
\caption{\textbf{Comparison to on-the-fly stylization with the training images.}}
\label{tab:retab2}
\end{table}
}

\begin{abstract}
\label{sec:abstract}
In generative adversarial networks, improving discriminators is one of the key components for generation performance. As image classifiers are biased toward texture and debiasing improves accuracy, we investigate 1) if the discriminators are biased, and 2) if debiasing the discriminators will improve generation performance. Indeed, we find empirical evidence that the discriminators are sensitive to the style (\eg, texture and color) of images. As a remedy, we propose \emph{feature statistics mixing regularization} (\emph{FSMR}) that encourages the discriminator's prediction to be invariant to the styles of input images. Specifically, we generate a mixed feature of an original and a reference image in the discriminator's feature space and we apply regularization so that the prediction for the mixed feature is consistent with the prediction for the original image. We conduct extensive experiments to demonstrate that our regularization leads to reduced sensitivity to style and consistently improves the performance of various GAN architectures on nine datasets. In addition, adding FSMR to recently-proposed augmentation-based GAN methods further improves image quality. Our code is available at \url{https://github.com/naver-ai/FSMR}.


\end{abstract}

\section{Introduction}
\label{sec:intro}
Generative adversarial networks (GANs) \cite{Goodfellow2014generative} have achieved significant development over the past several years, enabling many computer vision and graphics applications  \cite{zhu2017unpaired, huang2018multimodal, kim2019u, choi2018stargan, choi2020stargan, park2019semantic, kim2021stylemapgan, kim2019tag2pix}. On top of the carefully designed architectures \cite{miyato2018cgans, radford2015unsupervised, zhang2019self, karras2017progressive, karras2019style, karras2020analyzing, brock2018large}, GAN-specific data augmentation and regularization techniques 
have been keys for improvements. Regularization techniques \cite{jenni2019stabilizing, gulrajani2017improved, mescheder2018training, miyato2018spectral, zhang2019consistency, zhao2020improved, jeong2021training, kang2020contragan} stabilize the training dynamics by penalizing steep changes in the discriminator’s output within a local region of the input. On the other hand, data augmentation techniques \cite{karras2020training, zhao2020differentiable} prevent the discriminator from overfitting as commonly adopted in classification domains. 
Note that both efforts aim to guide the discriminator not to fixate on particular subsets of observations and to generalize over the entire data distribution.

Texture has been shown to provide a strong hint for classifiers \cite{geirhos2018imagenet, gatys2015texture, hermann2019origins}. If such a hint is sufficient enough to achieve high accuracy, the models tend not to learn the complexity of the intended task \cite{bahng2020learning}. 
As the GAN discriminators are inherently classifiers, we presume that they also tend to rely on textures to classify real and fake images. Accordingly, the generators would focus on synthesizing textures which are regarded as real by the biased discriminator. 
In this paper, we answer the two questions: 1) are discriminators sensitive to style (\eg, texture and color)? and 2) if yes, will debiasing the discriminators improve the generation performance?

To answer the first question, we define style distance as shown in \fref{fig:fig1}a. An ideal discriminator would produce small style distance because the two images have the same content. As we do not have a unit of measurement, we compute relative distance: the style distance divided by the content distance. In other words, we measure the sensitivity to style as multiples of the distance between images with different content. Surprisingly, \fref{fig:fig1}b shows that all baselines have noticeable values in relative distance.

To answer the second question, we debias the discriminators and measure improvements in generative performance.
A straightforward approach for debiasing is to suppress the difference in the discriminator's output with respect to the style changes of the input image. Indeed, we observe that imposing a consistency loss \cite{zhang2019consistency,zhao2020improved} on the discriminator between the original image and its stylized version improves the generator as mimicking contents becomes easier than mimicking style to fool the discriminator.

However, this approach leads to other difficulties: the criteria for choosing style images are unclear, and stylizing all training images with various style references requires a huge computational burden and an external style dataset.
To efficiently address the style bias issue, we propose \emph{feature statistics mixing regularization} (\emph{FSMR}) which encourages the discriminator's prediction to be invariant to the styles of input images by mixing feature statistics within the discriminator. Specifically, we generate mixed features by combining original and reference features in the discriminator’s intermediate layers and impose consistency between the predictions for the original and the mixed features. 


In the experiments, we show that FSMR indeed induces the discriminator to have reduced sensitivity to style (\sref{sec:straightforward}).
We then provide thorough comparisons to demonstrate that FSMR consistently improves various GAN methods on benchmark datasets (\sref{sec:standard_dataset}). 
Our method can be easily applied to any setting without burdensome preparation. Our implementation and models will be publicly available online for the research community. Our contributions can be summarized as follows:
\begin{itemize}
  \item To the best of our knowledge, our work is the first style bias analysis for the discriminator of GANs. 
  \item We define the relative distance metric to measure the sensitivity to the styles (\sref{sec:style_bias}). 
  \item We propose feature statistics mixing regularization (FSMR), which makes the discriminator’s prediction to be robust to style (\sref{sec:method}).
  \item FSMR does not use external style images and outperforms the straightforward solution with external style images (\sref{sec:straightforward}).
  \item FSMR improves five baselines on all standard and small datasets regarding FID and relative distance (\sref{sec:standard_dataset}, \ref{sec:small_dataset}).
\end{itemize}

\figsensitivity

\figmodel

\section{Style-bias in GANs}
\label{sec:style_bias}

Our work is motivated by the recent finding that CNNs are sensitive to style rather than content, \ie, ImageNet-trained CNNs are likely to make a style-biased decision when the style cue and content cue have conflict \cite{geirhos2018imagenet}. 
To quantitatively measure how sensitive a discriminator is to style, we compute style distance, content distance, and then relative distance. Afterward, we describe a straightforward baseline solution to reduce the discriminator's distance to style.

\subsection{Style distance and content distance}
\label{sec:distance}
We define a quantitative measure for how sensitive a discriminator is sensitive to style.
First, given a set of training images, we utilize a style transfer method to synthesize differently stylized images of the same content. 
The styles are randomly chosen from WikiArt \cite{wikiart}.
Figure \ref{fig:fig1}a shows some example stylized images from AFHQ \cite{choi2020stargan}.
We define style distance $d_s$ between images with different styles and the same content. The content distance $d_c$ is defined vice versa:
\begin{equation}
\label{eq:styledistance}
    \underbrace{d_s(\bmc, \bms_1, \bms_2)}_{\text{style \ distance}} = d(T(\bmc, \bms_1), T(\bmc, \bms_2)),
\end{equation}
\begin{equation}
\label{eq:contentdistance}
    \underbrace{d_c(\bms, \bmc_1, \bmc_2)}_{\text{content \ distance}} = d(T(\bmc_1, \bms), T(\bmc_2, \bms)),
\end{equation}
where $T(\bmc, \bms)$ transfers the style of the reference image $\bms$ $\in \mathds{R}^{C \times H \times W}$ to the content image $\bmc$ $\in \mathds{R}^{C \times H \times W}$, and $d$ measures cosine distance in the last feature vectors of the discriminator. In practice, we use adaptive instance normalization (AdaIN) \cite{huang2017arbitrary} as $T$. \fref{fig:fig1} illustrates the process of calculating the content and style distances in Eq. \eqref{eq:styledistance} and \eqref{eq:contentdistance}. 

As we do not have a unit of measurement, we compute relative distance $\rho$, \ie, the style distance divided by the content distance:
\begin{equation}
\label{eq:rho}
    \underbrace{\rho}_{\text{relative distance}}\hspace{-1mm}= \mathop{\mathbb{E}}_{\substack{\bmc_1, \bmc_2 \in \mathbf{C}, \\ \bms_1, \bms_2 \in \mathbf{S}}}\left[\frac{d_s(\bmc_1, \bms_1, \bms_2)}{d_c(\bms_1, \bmc_1, \bmc_2)}\right],
\end{equation}
where $\mathbf{C}$ and $\mathbf{S}$ denote the training dataset and an external style dataset, respectively.
The larger the $\rho$ value, the more sensitive the discriminator is to style when classifying real and fake images. 
We will use the relative distance $\rho$ for further analysis from here on.
Our goal is to reduce the style distance so that the discriminators consider contents more important and produce richer gradients to the generators.

The relative distances of ImageNet-pretrained ResNet50 and ResNet50 pretrained for classifying Stylized ImageNet \cite{geirhos2018imagenet} supports validity of the metric. As the relative distance of the latter is less than the former and the latter is proven to be less biased toward style, we argue that the discriminators with lower relative distance are less sensitive to style (figures are deferred to \sref{sec:standard_dataset}).

\subsection{Baseline: On-the-fly stylization}
\label{sec:stylizeddataset}
A well-known technique for preventing the classifiers from being biased toward styles is to augment the images with their style-transferred versions, especially using the WikiArt dataset~\cite{wikiart} as style references~\cite{geirhos2018imagenet}.
It works because the style transfer does not alter the semantics of the original images or the anticipated output of the network.
On the other hand, in GAN training, style transfer drives the images out of the original data distribution, thus it changes the anticipated output of the discriminator \cite{karras2020training}. There are two workarounds for such a pitfall: 1) applying stochastic augmentations for both real and fake data \cite{karras2020training, zhao2020differentiable} and 2) penalizing the output difference caused by the augmentation instead of feeding the augmented images to the discriminator \cite{zhang2019consistency, zhao2020improved}.
As our goal is to make the discriminator less sensitive to style changes, we take the second approach as a straightforward baseline, for example, imposing consistency on the discriminator between the original images $\bmc$ and their randomly stylized images $T(\bmc, \bms)$ by 
\begin{equation}
\label{eq:ontheflyreg}
    L_{\text{consistency}}=\mathbb{E}_{\bmc, \bms}\left[(\bmD(\bmc)-\bmD(T(\bmc, \bms)))^2\right],
\end{equation}
where $D(\centerdot)$ denotes the logit from the discriminator.
However, it raises other questions and difficulties: the criteria for choosing the style images are unclear, and stylizing each image on-the-fly requires additional costs and an external dataset. Another option is to prepare a stylized dataset instead of on-the-fly stylization but it further requires prohibitively large storage. To combat this, we propose an efficient and generally effective method, feature mixing statistics regularization, whose details are described in the next section.

\section{Proposed method}
\label{sec:method}
We first describe the traditional style transfer algorithm, AdaIN, as a preliminary. Then, we discuss how our proposed method, feature statistics mixing regularization (FSMR), incorporates AdaIN to induce the discriminator to be less sensitive to style.

\subsection{Preliminary: AdaIN}
\label{sec:preliminary}

Instance normalization (IN) \cite{ulyanov2016instance} performs a form of style removal by normalizing feature statistics. Adaptive instance normalization (AdaIN) \cite{huang2017arbitrary} extends IN to remove the existing style from the content image and transfer a given style. Specifically, AdaIN transforms content feature maps $\bmhx$ into feature maps whose channel-wise mean and variance are the same as those of style feature maps $\bmhxt$:

\begin{equation}
\label{eq:adain}
    \adain({\bmhx}, {\bmhxt}) = \bmsigma({\bmhxt})\left(\frac{{\bmhx} - \bmmu({\bmhx})}{\bmsigma(\bmhx)}\right) + \bmmu({\bmhxt}),
\end{equation}
where $\bmhx, \bmhxt \in \real^{C\times H\times W}$ are features obtained by a pre-trained encoder, and $\bmmu(\cdot)$ and $\bmsigma(\cdot)$ denote their mean and standard deviation their spatial dimensions, calculated for each channel, respectively. 
Then, through a properly trained decoder, the transformed features become a stylized image\footnote{AdaIN may denote the full stylization process but it denotes the operation on the feature maps (\eref{eq:adain}) in this paper.}.
Much work has adopted AdaIN within the \textit{generator} for improving the generation performance \cite{huang2018multimodal,kim2019u, karras2019style, choi2020stargan, kim2021stylemapgan, kim2019tag2pix}. On the contrary, our proposed method (FSMR) employs it within the \textit{discriminator} for efficient regularization, as described below.

\subsection{Feature statistics mixing regularization}
\label{sec:fsmr}
Our goal is to make the discriminator do not heavily rely on the styles of the input images, without suffering from the difficulties of the on-the-fly stylization (\sref{sec:stylizeddataset}).
Hence, we propose feature statistics mixing regularization (FSMR), which does not require any external dataset and can be efficiently implemented as per-layer operations in the discriminator.
FSMR mixes the mean and standard deviation of the intermediate feature maps in the discriminator using another training sample and penalizes discrepancy between the original output and the mixed one.

Specifically, we define feature statistics mixing (FSM) for feature maps $\bmhx$ with respect to feature maps $\bmhxt$ to be AdaIN followed by linear interpolation:

\begin{equation}
\label{eq:fsm}
\begin{aligned}
    \fsm(\bmhx, \bmhxt) = \alpha\bmhx+(1-\alpha)\adain(\bmhx, \bmhxt),
\end{aligned}
\end{equation}
where $\alpha\sim \text{Uniform}(0,1)$ controls the intensity of feature perturbation. We suppose that varying $\alpha$ will let the discriminator learn from various strengths of regularization.

Denoting an $i$-th layer of the discriminator as $\bmf_{i}$, a content image as $\bmc$, and a style reference image as $\bms$ which is randomly chosen from the current mini-batch samples, we define the mixed feature maps $\tilde{\bmx}$ and $\tilde{\bmy}$ through feed-forward operations with FSM:
\begin{equation}
\label{eq:forward}
\begin{aligned}
    & \tilde{\bmx}_1 = \bmx_1 = \bmf_{1}(\bmc), \\
    & \tilde{\bmy}_1 = \bmy_1 =\bmf_{1}(\bms), \\
    & \tilde{\bmx}_{i+1} = \bmf_{i+1}(\fsm(\tilde{\bmx}_{i}, \tilde{\bmy}_{i})), \\
    & \tilde{\bmy}_{i+1} = \bmf_{i+1}(\fsm(\tilde{\bmy}_{i}, \tilde{\bmx}_{i})). \\
\end{aligned}
\end{equation}
Then the final output logit of the mixed feed-forward pass through the discriminator with $n$ convolutional layers becomes:
\begin{equation}
\label{eq:logit}
\begin{aligned}
    & \bmD_\fsm(\bmc, \bms)=\linear(\tilde{\bmx}_n).
\end{aligned}
\end{equation}
Given the original output $\bmD(\bmc)$ and the mixed output $\bmD_{\text{FSM}}(\bmc, \bms)$, we penalize their discrepancy with a loss: 
\begin{equation}
\label{eq:loss}
\begin{aligned}
    & L_{\text{FSMR}}=\mathbb{E}_{\bmc, \bms \sim p_{\text{data}}}\left[(\bmD(\bmc)-\bmD_\fsm(\bmc, \bms))^2\right].
\end{aligned}
\end{equation}

\fref{fig:fig2} illustrates the full diagram of FSMR. This loss is added to the adversarial loss~\cite{Goodfellow2014generative} when updating the discriminator parameters. It regularizes the discriminator to produce consistent output under different statistics of the features varying through the layers. Our design of $L_{\text{FSMR}}$ is general-purpose and thereby can be combined with other methods~\cite{karras2019style, zhao2020differentiable, karras2020training}. 
As shown in Algorithm \ref{alg:fsmcode}, FSM can be implemented with only a few lines of code. Also, we provide the Tensorflow-like pseudo-code of FSMR in Appendix \ref{app:pseudo}.

\begin{algorithm}[t]
\caption{FSM Pseudocode, Tensorflow-like}
\label{alg:fsmcode}
\definecolor{codeblue}{rgb}{0.25,0.5,0.5}
\definecolor{codekw}{rgb}{0.85, 0.18, 0.50}
\lstset{
  backgroundcolor=\color{white},
  basicstyle=\fontsize{7.5pt}{7.5pt}\ttfamily\selectfont,
  columns=fullflexible,
  breaklines=true,
  captionpos=b,
  commentstyle=\fontsize{7.5pt}{7.5pt}\color{codeblue},
  keywordstyle=\fontsize{7.5pt}{7.5pt}\color{codekw},
}
\begin{lstlisting}[language=python]
# N: batch size, H: height, W: width, C: channels
def FSM(x, y, eps=1e-5):
    x_mu, x_var = tf.nn.moments(x, axes=[1,2])
    y_mu, y_var = tf.nn.moments(y, axes=[1,2])
    
    # normalize
    x_norm = (x - mu) / tf.sqrt(var + eps)
    
    # de-normalize
    x_fsm = x_norm * tf.sqrt(y_var + eps) + y_mu
    
    # combine
    alpha = tf.random.uniform(shape=[])
    
    x_mix = alpha * x + (1 - alpha) * x_fsm
    
    return x_mix # NxHxWxC
\end{lstlisting}
\end{algorithm}
\figrecon

\subsection{Visualizing the effect of FSM}
\label{sec:visualization}
To visually inspect the effect of FSM in the discriminator, we train a decoder (same architecture as the one for AdaIN \cite{huang2017arbitrary}) which reconstructs the original image from the $32\times 32$ feature maps of the \textit{original} discriminator.

In \fref{fig:fig3}, the content images go through the discriminator with FSM on all layers with respect to the style images to produce stylized (\ie, \textit{FSMed}) intermediate features. Then the learned decoder synthesizes the result images from the FSMed features.

The FSMed images have similar global styles to the style images but contain semantics of the content images.
It has a similar effect to AdaIN but better preserves the fine details of the content. We suggest that it is the key for the discriminator to be able to provide gradients toward more realistic images for the generator leading to higher quality images than the on-the-fly stylization baseline (\sref{sec:straightforward}).


\figensitivityresult

\section{Experiments}
\label{sec:experiments} 
We conduct extensive experiments on six datasets of CIFAR-10\cite{krizhevsky2009learning}, FFHQ\cite{karras2019style}, AFHQ\cite{choi2020stargan}, CelebA-HQ\cite{karras2017progressive}, LSUN Church\cite{yu15lsun}, and MetFaces\cite{karras2020training} with five GAN methods such as DCGAN\cite{radford2015unsupervised}, bCRGAN\cite{zhao2020improved},
StyleGAN2\cite{karras2020analyzing}, DiffAugment\cite{zhao2020differentiable}, and ADA\cite{karras2020training}. We choose the datasets and baseline methods following the recent experimental setups \cite{zhao2020differentiable,karras2020training}. We use the relative distance $\rho$ (\eref{eq:rho}), Fr\'echet inception distance (FID) \cite{heusel2017gans}, and inception score (IS) \cite{salimans2016improved} as evaluation metrics. When we compute FID, we use all training samples and the same number of fake samples. All the baseline methods are trained using the official implementations provided by the authors. See Appendix \ref{app:details} for more details.
We next conduct thorough experiments to demonstrate the superiority of our method over the straightforward solution and the baselines.

\tabfidgan
\tabstytime

\subsection{Comparison with the on-the-fly stylization}
\label{sec:straightforward}
In this section, we compare our method with the on-the-fly stylization, \ie, generating stylized images via AdaIN during training and applying consistency regularization (Section \ref{sec:stylizeddataset}). To perform this, we collect 100 style images from WikiArt~\cite{wikiart} and randomly sample one for stylizing each image during training. Note that, unlike the on-the-fly stylization, FSMR does not rely on external style images.
We conduct experiments on five benchmark datasets: CIFAR-10, CelebA-HQ, FFHQ, AFHQ, and LSUN Church.

\tref{tab:table1} compares effect of regularization with on-the-fly stylization and FSMR in FID. While the former improves FID compared to the baselines to some extent, improvements due to FSMR are larger in all cases. For comparison with additional networks and datasets, see Appendix \ref{app:addresults}. 

To measure the discriminator's sensitivity to style, we compute the relative distance $\rho$ (\eref{eq:rho}) for each method. \fref{fig:fig4} shows the relative distance on CIFAR-10, FFHQ, and AFHQ. As one can easily expect, utilizing the stylized dataset reduces the discriminator's sensitivity toward style. It is worth noting that FSMR not only consistently reduces the sensitivity but also outperforms the competitor in all cases.
This is a very meaningful result because FSMR \textit{does not use any external stylized dataset} but it uses only the original images during training. We also observe that the lower relative distances agree with the lower FIDs within the same environment.

We compare the time and memory costs in \tref{tab:table1}. FSMR requires 3.0$\sim$7.4\% extra training time, but the on-the-fly method requires 17.2$\sim$26.8\% extra training time for additional feedforward passes in image stylization. In addition, the on-the-fly method requires 70.0$\sim$87.5\% extra GPU memory to hold pretrained networks and features for image stylization, but FSMR only adds negligible ($\sim$2\%) GPU memory.
To avoid extra costs for the on-the-fly stylization during training, we can prepare the stylized datasets before training (\ie, different approach but has the same effect as the on-the-fly stylization).
However, the one-to-many stylization in advance requires heavy computation and prohibitively large storage as shown in \tref{tab:tabstytime}. For example, to construct the stylized dataset for 1024$\times$1024 FFHQ with 100 style references, we need to process and store more than 7.0M (70k $\times$ 100) images (8.93TB).

As an ablation study, we push toward harsher regularization: using randomly shifted feature maps instead of FSM. We observe that using arbitrary mean and standard deviation in $\text{AdaIN}$ (\eref{eq:adain}) significantly hampers adversarial training between a generator and a discriminator, \ie, the training diverges. On the other hand, FSMR using in-domain samples shows the anticipated effect.


\tabfidstylegan 
\figCIFAR

\subsection{Standard datasets}
\label{sec:standard_dataset}

We evaluate the effectiveness of FSMR on three benchmark datasets, all of which have more than 10k training images: CIFAR-10 (50k), FFHQ (70k), and AFHQ (16k). \tref{tab:table2} (left) shows that FSMR consistently improves StyleGAN2 even with existing augmentation techniques \cite{karras2020training, zhao2020differentiable}. We emphasize that FSMR enhances baselines by a large gap on AFHQ, in which case the discriminator might be easily biased toward color and texture of the animals. 

\fref{fig:fig5} shows the relative distances on CIFAR-10, FFHQ, and AFHQ for StyleGAN2 variants. FSMR reduces the relative distances in all cases and they agree with the improvements in FID. We also provide the relative distances of ResNet50 networks pretrained on ImageNet and Stylized ImageNet as references in each dataset (\sref{sec:distance}). As the lower relative distances agree with the higher classification performances, the lower relative distances of the discriminator agree with the higher generative performances.

In addition, \tref{tab:table3} demonstrates that applying FSMR on StyleGAN2 variants further improves both FID and IS for both unconditional and class-conditional generation on CIFAR-10. For qualitative results, see \fref{fig:fig6} and Appendix \ref{app:addresults}.

\tabCIFAR
\figresultexample

\subsection{Small datasets.}
\label{sec:small_dataset}
GANs are known to be notoriously difficult to train on small datasets due to limited coverage of the data manifold. Being able to train GANs on small datasets would lead to a variety of application domains, making a rich synthesis experience for the users.
We tried our method with five small datasets that consist of a limited number of training images such as MetFaces (1k), AFHQ Dog (5k), AFHQ Cat (5k). AFHQ Wild (5k). As shown in Table \ref{tab:table2} (right), we can observe that FSMR improves FID stably for all the baseline models, even if the number of data is small. See \fref{fig:fig6} and Appendix \ref{app:addresults} for qualitative results.

\section{Related Work}
\label{sec:related}

\vspace{-1mm}
\paragraph{Improving discriminators.}
While generative adversarial networks \cite{Goodfellow2014generative} have been developing regarding their network architectures \cite{radford2015unsupervised, mescheder2018training, karras2019style, karras2020analyzing}, regularizing the discriminator has been simultaneously considered as an important technique for stabilizing their adversarial training. Examples include instance noise\cite{jenni2019stabilizing}, gradient penalties\cite{gulrajani2017improved, mescheder2018training}, spectral normalization\cite{miyato2018spectral}, contrastive learning\cite{jeong2021training, kang2020contragan}, and consistency regularization\cite{zhang2019consistency, zhao2020improved}. They implicitly or explicitly enforce smooth changes in the outputs within some perturbation of the inputs. Recent methods employ data augmentation techniques to prevent discriminator overfitting \cite{karras2020training, zhao2020differentiable}. While they explicitly augment the images, our method implicitly augments the feature maps in the discriminator. In addition, while they use standard transformations which are used in training classifiers, our method regularizes the discriminator to produce small changes when the style of the input image is changed and it effectively prevents discriminator from being biased toward style.

\paragraph{Bias toward style.}
Convolutional neural networks are biased toward style (texture) when they are trained for classification \cite{gatys2015texture,geirhos2018imagenet, hermann2019origins}. The straightforward solution for reducing the bias is randomizing textures of the samples by a style transfer algorithm \cite{geirhos2018imagenet}. It is a kind of data augmentation technique in that the style transfer prevents classifiers from overfitting to styles as geometric or color transformations prevent classifiers from overfitting to certain positions or colors. As simply perturbing the data distribution in GAN training results in perturbed fake distribution \cite{karras2020training}, we introduce an extra forward pass with an implicitly stylized feature and impose consistency in the output with respect to the original forward pass (\eref{eq:loss}). While the linear interpolation of our mixing resembles mixup \cite{zhang2017mixup}, we do not interpolate target outputs but only soften the changes in feature statistics.

\paragraph{Style mixing regularization}~\cite{karras2019style} may look similar to FSMR in that it also mixes two styles. It mixes styles in the generator and encourages the generator to produce mixed images that will be used in the adversarial training for both the generator and the discriminator. Its goal is to divide the role of layers and it has little effect on performance (4.42$\rightarrow$4.40, FFHQ, StyleGAN, 1024x1024 resolution). On the other hand, FSMR implicitly mixes styles in the discriminator and suppresses sensitivity to style by imposing consistency regularization to the discriminator. FSMR has a great influence on performance improvement (5.52$\rightarrow$3.72, FFHQ, StyleGAN2, 256x256 resolution).

\section{Limitation and Discussion}
\label{sec:limit}

As shown in various experiments, we have found that the discriminators have a bias for style, which enables numerical representation through the relative distance metric. However, we have not found out the optimal value that how much relative distance should be reduced for each model. We observed through the reference value in \fref{fig:fig5}, that even though we could not find the optimal value, the relationship where the relative distance decreases, the less bias to style reduces. We have proposed FSMR, which reduces the bias to style using only internal training datasets, rather than using external datasets, and proved that FSMR is very simple yet effective. In future work, it would be worthwhile to search the optimal value for the relative distances and to unify the relative distances among different models.

\section{Conclusion}
\label{sec:conclusion}

We observed that the discriminators are biased toward style. To quantitatively measure the amount of bias, we proposed relative distance, \ie, style distance divided by the content distance. While reducing the style bias with a straightforward consistency regularization with style transfer method induces ambiguity and difficulties, our feature statistics mixing regularization (\textit{FSMR}) provides a simple and effective solution. Importantly, FSMR does not explicitly stylize the images but perturbs the intermediate features in the discriminator. We visualize the effect of FSMR and quantitatively analyze its behavior regarding relative sensitivity. 
The experiments demonstrated that our method consistently improves various network architectures, even in conjunction with the latest techniques. 

\vspace{-1mm}
\paragraph{Acknowledgements} The authors thank NAVER AI Lab researchers and Jun-Yan Zhu for constructive discussion. All experiments were conducted on NAVER Smart Machine Learning (NSML) platform~\cite{kim2018nsml, sung2017nsml}. This work was partly supported by an IITP grant (No.2021-0-00155) and an NRF grant (NRF-2021R1G1A1095637). Both grants are funded by the Korean government (MSIT).

\newpage
{\small
\bibliographystyle{ieee_fullname}
\bibliography{egbib}
}

\newpage
\clearpage
\begin{appendix}

\section{Implementation Details}
\label{app:details}

Figure \ref{fig:figsupstyexam} presents all examples of style images which are used in computing relative distance and running the baseline on-the-fly stylization.
In the subsections, we provide implementation details for DCGAN variants and StyleGAN2 variants. For further details, please refer to our code at \url{https://github.com/naver-ai/FSMR}.

\subsection{DCGAN variants Experiments}
Augmentations for DCGAN, bCRGAN include image flipping and random cropping. Consistency regularization coefficients for bCRGAN are $\lambda_{real}=\lambda_{fake}=10$.
We use non-saturating logistic loss with $R_1$ regularization, and Adam optimizer with $\beta_1=0.5$, $\beta_2=0.999$, $\epsilon=10^{-8}$, and learning rate=0.0001 for both models. We ran our experiments on one Tesla V100 GPU, using Tensorflow 2.1.0, CUDA 10.1, and cuDNN 7.6.4. 
We apply FSMR for both real and fake samples
and the total loss is
\begin{equation}
\label{eq:loss}
    L_{\text{total}} = L_{\text{adv}} + \lambda L_{\text{FSMR}},
\end{equation}
where $\lambda=10$. We performed all training runs using 1 GPU, continued the training for 20k iterations, and used a minibatch size of 32.

\subsection{StyleGAN2 variants Experiments}
For StyleGAN2, ADA\footnote{\url{https://github.com/NVlabs/stylegan2-ada}}, and DiffAug\footnote{\url{https://github.com/mit-han-lab/data-efficient-gans}}, we use the official Tensorflow implementations. We kept most of the details unchanged, including network architectures, weight demodulation, path length regularization, lazy regularization, style mixing regularization, bilinear filtering in all up/downsampling layers, equalized learning rate for all trainable parameters, minibatch standard deviation layer at the end of the discriminator, exponential moving average of generator weights, non-saturating logistic loss with $R_1$ regularization, and Adam optimizer with $\beta_1=0$, $\beta_2=0.99$, and $\epsilon=10^{-8}$. We ran our experiments on 
eight Tesla V100 GPUs, using Tensorflow 1.14.0, CUDA 10.0, and cuDNN 7.6.3.
We apply FSMR only to the real samples because it leads to slightly larger gain.
The weights $\lambda$ for $L_{\text{FSMR}}$ are 0.05 for FFHQ and 1 for the other datasets. We performed all training runs using 4 GPUs, continued the training for 25M iterations, and used minibatch size of 32, except for CIFAR-10, where we used 2 GPUs, 100M iterations, and a minibatch size of 64.

\section{Evaluation metrics}
\label{app:evalmetric}
We measure Fr\'echet Inception Distance (FID) \cite{heusel2017gans} and Inception Score (IS) \cite{salimans2016improved} using the official Inception v3 model in Tensorflow. When we compute FID and IS, we sample the same number of images to the number of real images in the training set. 




\section{Pseudo-code}
\label{app:pseudo}
We provide the Tensorflow-like pseudo-code of FSMR in Algorithm \ref{alg:code}.
FSMR is simple to fit easily into any model.

\begin{algorithm}[t]
\caption{FSMR Pseudocode, Tensorflow-like}
\label{alg:code}
\definecolor{codeblue}{rgb}{0.25,0.5,0.5}
\definecolor{codekw}{rgb}{0.85, 0.18, 0.50}
\lstset{
  backgroundcolor=\color{white},
  basicstyle=\fontsize{7.5pt}{7.5pt}\ttfamily\selectfont,
  columns=fullflexible,
  breaklines=true,
  captionpos=b,
  commentstyle=\fontsize{7.5pt}{7.5pt}\color{codeblue},
  keywordstyle=\fontsize{7.5pt}{7.5pt}\color{codekw},
}
\begin{lstlisting}[language=python]
# N: batch size, H: height, W: width, C: channels
def FSM(x, y, alpha, eps=1e-5):
    x_mu, x_var = tf.nn.moments(x, axes=[1,2], keepdims=True) # Nx1x1xC
    y_mu, y_var = tf.nn.moments(y, axes=[1,2], keepdims=True) # Nx1x1xC
    
    # normalize
    x_norm = (x - mu) / tf.sqrt(var + eps)
    
    # de-normalize
    x_fsm = x_norm * tf.sqrt(y_var + eps) + y_mu
    
    # combine
    x_mix = alpha * x + (1 - alpha) * x_fsm
    
    return x_mix # NxHxWxC

def discriminator(img, use_fsmr=False):
    # layers: conv, bn, actv, ..., fc -> discriminator layers
    
    x = img # NxHxWxC
    indices = tf.range(tf.shape(x)[0])
    shuffle_indices = tf.random.shuffle(indices)
    alpha = tf.random.uniform(shape=[], minval=0.0, maxval=1.0)
    
    for layer in layers:
        x = layer(x)
        if use_fsmr and layer.name == 'conv':
            y = tf.gather(x, shuffle_indices)
            x = FSM(x, y, alpha)
        
    return x # Nx1
	
def FSMR(real_img, fake_img, use_fsmr=True):
    real_logit = discriminator(real_img) # Nx1
    fake_logit = discriminator(fake_img) # Nx1
    
    if use_fsmr:
    	real_logit_mix = discriminator(real_img, use_fsmr) # Nx1
    	fake_logit_mix = discriminator(fake_img, use_fsmr) # Nx1
    	d_fsmr_loss = l2_loss(real_logit, real_logit_mix)
    	d_fsmr_loss += l2_loss(fake_logit, fake_logit_mix) # optional
    	d_fsmr_loss *= 10 # weight for fsmr
    else:
        d_fsmr_loss = 0

    return d_fsmr_loss

    
\end{lstlisting}
\end{algorithm}

\section{Comparison with previous mixing methods.}
\label{app:compresults}

In Table \ref{tab:retab1}, we show the comparison results from the previous mixing methods. First of all, the difference between the previous methods and FSMR is as follows. \emph{CutMix}\cite{yun2019cutmix} and \emph{Mixup}\cite{zhang2017mixup} apply augmentations on images, not on the feature maps. \emph{Manifold-Mixup}\cite{verma2019manifold} performs linear interpolation on the feature map without implicit style transfer. \emph{StyleMix}\cite{hong2021stylemix} applies AdaIN on images with an extra encoder for augmentation, not for consistency. \emph{MoEx}\cite{li2021feature} shares only the first component with ours: reducing style bias requires not only feature statistics mixing (e.g. AdaIN\cite{huang2017arbitrary}, MoEx), but also the consistency loss after feature statistics mixing.
When we have conducted the experiment, CutMix and Manifold-Mixup are applied as follows: CutMix in feature maps and Manifold-Mixup with consistency regularization. None of them outperforms ours because they do not reduce style bias.

\retabmix

\section{Ablation on the style dataset.}
\label{app:styledataset}
\tref{tab:retab2} shows that the improvements by FSMR is larger than the improvements by using internal images as style references. The numbers (a) / (b) / (c) report FIDs for
\begin{enumerate}[label=(\alph*),itemsep=-1mm]
    \item on-the-fly with the training images
    \item on-the-fly with WikiArt
    \item FSMR.
\end{enumerate}
FSMR outperforms both on-the-fly settings. Hence, we argue that the difference does not come from using the same data distribution but from FSMR.

\retabfly

\section{Additional results}
\label{app:addresults}

In \fref{fig:figsupgan}, we show the relative distance on several datasets except the ones in the main paper. We observe that FSMR reduces the relative distance of the discriminator, \ie, the discriminator relies less on style. 


Figure \ref{fig:figreconsup} illustrate an additional visualization of the effect of FSM on FFHQ, AFHQ, and LSUN Church. 

In Figure \ref{fig:figsampleresult}, and \ref{fig:figuncurateall}, we show generated images for several datasets from the baseline with FSMR. The images were selected at random, \ie, we did not perform any cherry-picking. We observe that FSMR yields excellent results in all cases.





\newpage

\figsupstyexam{}
\figsupgan{}


\figreconsup{}

\figsampleresult{}
\figuncurateall{}









\end{appendix}

\end{document}